%% file: main.tex
\documentclass{article}

\usepackage[final]{neurips_2020}

\usepackage[utf8]{inputenc} 
\usepackage[T1]{fontenc}    
\usepackage{url}            
\usepackage{booktabs}       
\usepackage{amsfonts}       
\usepackage{nicefrac}       
\usepackage{microtype}      

\usepackage{xcolor}
\definecolor{darkred}{rgb}{0.4,0.0,0.0}
\definecolor{darkgreen}{rgb}{0.0,0.4,0.0}
\definecolor{darkblue}{rgb}{0.0,0.0,0.4}
\usepackage{hyperref}
\hypersetup{colorlinks, citecolor=darkblue, urlcolor=darkblue}

\usepackage{graphicx}
\usepackage{subcaption}
\usepackage{wrapfig}
\usepackage[most]{tcolorbox}

\graphicspath{{figures/illustrations/}}
\newcommand\sampleswidth{0.49\linewidth} 

\usepackage{multirow}

\usepackage{algorithm}
\usepackage{algorithmic}

\usepackage{amsmath}
\usepackage{amssymb}

\input{defns}

\title{Closing the Dequantization Gap: \\PixelCNN as a Single-Layer Flow}

\author{
  Didrik Nielsen \\
  Technical University of Denmark \\
  \texttt{didni@dtu.dk}
  \And
   Ole Winther \\
   Technical University of Denmark \\
  \texttt{olwi@dtu.dk}
}

\begin{document}

\maketitle

\begin{abstract}
    Flow models have recently made great progress at modeling ordinal discrete data such as images and audio. Due to the continuous nature of flow models, dequantization is typically applied when using them for such discrete data, resulting in lower bound estimates of the likelihood. 
    In this paper, we introduce \emph{subset flows}, a class of flows that can tractably transform finite volumes and thus allow \emph{exact} computation of likelihoods for discrete data. 
    Based on subset flows, we identify ordinal discrete autoregressive models, including WaveNets, PixelCNNs and Transformers, as single-layer flows.
    We use the flow formulation to compare models trained and evaluated with either the exact likelihood or its dequantization lower bound. 
    Finally, we study multilayer flows composed of PixelCNNs and non-autoregressive coupling layers and demonstrate state-of-the-art results on CIFAR-10 for flow models trained with dequantization.
    
\end{abstract}

\section{Introduction}
\input{_introduction}

\section{Background}
\label{sec:background}
\input{_background}

\section{Closing the Dequantization Gap}
\label{sec:dequantization_gap}
\input{_subset_flows}

\section{PixelCNN as a Single-Layer Flow}
\input{_pixelcnn_as_flow}

\section{Related Work}
\input{_related_work}

\section{Experiments}
\input{_experiments}

\section{Conclusion}
\input{_conclusion}

\section*{Acknowledgements}
\input{_acknowledgements}

\section*{Funding Disclosure}
\input{_funding_disclosure}

\section*{Broader Impact}
\input{_broader_impact}

\bibliography{references}
\bibliographystyle{apalike}

\newpage
\onecolumn
\appendix
\input{_appendix}

\end{document}

%% file: defns.tex
\newcommand{\vh}{\boldsymbol{h}}

\newcommand{\vr}{\boldsymbol{r}}
\newcommand{\vs}{\boldsymbol{s}}

\newcommand{\vu}{\boldsymbol{u}}
\newcommand{\vv}{\boldsymbol{v}}
\newcommand{\vw}{\boldsymbol{w}}
\newcommand{\vx}{\boldsymbol{x}}
\newcommand{\vy}{\boldsymbol{y}}
\newcommand{\vz}{\boldsymbol{z}}

\newcommand{\vtheta}{\boldsymbol{\theta}}

\newcommand{\vlambda}{\boldsymbol{\lambda}}
\newcommand{\vmu}{\boldsymbol{\mu}}

\newcommand{\vpi}{\boldsymbol{\pi}}

\newcommand{\E}{\mathbb{E}}
\newcommand{\I}{\mathbb{I}}
\DeclareMathOperator*{\argmax}{arg\,max}


%% file: _introduction.tex

\begin{wrapfigure}{r}{0.46\linewidth}
    \centering
    \vspace{-15pt}
    \setlength{\belowcaptionskip}{-10pt}
    \resizebox{1.0\linewidth}{!}{
    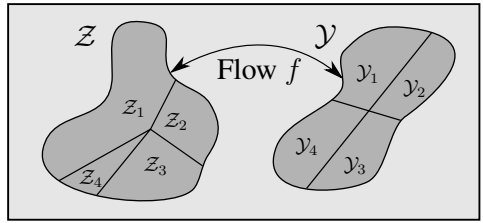
    }
    \caption{Subset flows $f: \mathcal{Y} \rightarrow \mathcal{Z}$ allow not only to transform points $\vz = f(\vy)$, but also subsets $\mathcal{Z}_i = f(\mathcal{Y}_i)$, in one pass. As a result, these flows can be trained on ordinal discrete data without the need for dequantization.}
    \label{fig:subset_flows}
\end{wrapfigure}

Learning generative models of high-dimensional data poses a significant challenge. The model will have to capture not only the marginal distributions of each of the variables, but also the potentially combinatorial number of interactions between them.
Deep generative models provide tools for learning richly-structured, high-dimensional distributions, utilizing the vast amounts of unlabeled data available. Generative adversarial networks (GANs) \citep{goodfellow2014} are one class of deep generative models that have demonstrated an impressive ability to generate plausible-looking images. However, GANs typically lack support over the full data distribution and provide no quantitative measure of performance. 
\emph{Likelihood-based deep generative models}, on the other hand, do provide this and can be classified as:
\begin{enumerate}
    \setlength\itemsep{0mm}
    \item \textbf{Latent variable models} such as Deep Belief Networks \citep{hinton2006, hinton2007}, Deep Boltzmann Machines \citep{salakhutdinov2009}, Variational Autoencoders (VAEs) \citep{kingma2013, rezende2014}.
    \item \textbf{Autoregressive models} such as Recurrent Neural Networks (RNNs), MADE \citep{germain2015}, WaveNet \citep{oord2016wavenet}, PixelCNN \citep{oord2016pixelrnn}, PixelCNN++ \citep{salimans2017}, Sparse Transformers \citep{child2019}.
    \item \textbf{Flow models} such as RealNVP \citep{dinh2017}, Glow \citep{kingma2018}, MAF \citep{papamakarios2017}, FFJORD \citep{grathwohl2018}.
\end{enumerate}
Data recorded from sensors are quantized before storage, resulting in ordinal data, i.e. discrete data with a natural ordering.
Autoregressive models excel at modelling such data since they can directly model discrete distributions. 
Apart from discrete flows \citep{tran2019,hoogeboom2019} -- which are severely restricted in expressiveness -- the vast majority of flow models are continuous and therefore require \emph{dequantization} to be applied to discrete data. However, dequantization comes at the cost of lower bound estimates of the discrete likelihood \citep{theis2016,ho2019}.

In this paper: 
\emph{1)} We introduce subset flows, a class of flows that allow tractable transformation of finite volumes and consequently may be trained directly on discrete data such as images, audio and video without the need for dequantization.
\emph{2)} Based on subset flows, we formulate existing autoregressive models for ordinal discrete data, such as PixelCNN \citep{oord2016pixelrnn} and PixelCNN++ \citep{salimans2017}, as single-layer autoregressive flows. 
\emph{3)} Using the flow formulation of PixelCNNs, we quantify how dequantization used in training and evalutation impacts performance.
\emph{4)} We construct multilayer flows using compositions of PixelCNNs and coupling layers \citep{dinh2017}. For CIFAR-10, we demonstrate state-of-the-art results for flow models trained with dequantization. The code used for experiments is publicly available at \url{https://github.com/didriknielsen/pixelcnn_flow}.

%% file: figures/illustrations/subset_flows.pdf_tex
\begingroup%
  \makeatletter%
  \providecommand\color[2][]{%
    \errmessage{(Inkscape) Color is used for the text in Inkscape, but the package 'color.sty' is not loaded}%
    \renewcommand\color[2][]{}%
  }%
  \providecommand\transparent[1]{%
    \errmessage{(Inkscape) Transparency is used (non-zero) for the text in Inkscape, but the package 'transparent.sty' is not loaded}%
    \renewcommand\transparent[1]{}%
  }%
  \providecommand\rotatebox[2]{#2}%
  \newcommand*\fsize{\dimexpr\f@size pt\relax}%
  \newcommand*\lineheight[1]{\fontsize{\fsize}{#1\fsize}\selectfont}%
  \ifx\svgwidth\undefined%
    \setlength{\unitlength}{228.18897638bp}%
    \ifx\svgscale\undefined%
      \relax%
    \else%
      \setlength{\unitlength}{\unitlength * \real{\svgscale}}%
    \fi%
  \else%
    \setlength{\unitlength}{\svgwidth}%
  \fi%
  \global\let\svgwidth\undefined%
  \global\let\svgscale\undefined%
  \makeatother%
  \begin{picture}(1,0.46013281)%
    \lineheight{1}%
    \setlength\tabcolsep{0pt}%
    \put(0,0){\includegraphics[width=\unitlength,page=1]{subset_flows.pdf}}%
    \put(0.24194614,0.22485763){\color[rgb]{0,0,0}\makebox(0,0)[lt]{\lineheight{0}\smash{\begin{tabular}[t]{l}$\mathcal{Z}_1$\end{tabular}}}}%
    \put(0.14028057,0.3690888){\color[rgb]{0,0,0}\makebox(0,0)[lt]{\lineheight{0}\smash{\begin{tabular}[t]{l}{\Large $\mathcal{Z}$}\end{tabular}}}}%
    \put(0.6497522,0.37095519){\color[rgb]{0,0,0}\makebox(0,0)[lt]{\lineheight{0}\smash{\begin{tabular}[t]{l}{\Large $\mathcal{Y}$}\end{tabular}}}}%
    \put(0,0){\includegraphics[width=\unitlength,page=2]{subset_flows.pdf}}%
    \put(0.33097736,0.19912496){\color[rgb]{0,0,0}\makebox(0,0)[lt]{\lineheight{0}\smash{\begin{tabular}[t]{l}$\mathcal{Z}_2$\end{tabular}}}}%
    \put(0.28847991,0.10416952){\color[rgb]{0,0,0}\makebox(0,0)[lt]{\lineheight{0}\smash{\begin{tabular}[t]{l}$\mathcal{Z}_3$\end{tabular}}}}%
    \put(0.15151036,0.0770487){\color[rgb]{0,0,0}\makebox(0,0)[lt]{\lineheight{0}\smash{\begin{tabular}[t]{l}$\mathcal{Z}_4$\end{tabular}}}}%
    \put(0,0){\includegraphics[width=\unitlength,page=3]{subset_flows.pdf}}%
    \put(0.73816908,0.29873511){\color[rgb]{0,0,0}\makebox(0,0)[lt]{\lineheight{0}\smash{\begin{tabular}[t]{l}$\mathcal{Y}_1$\end{tabular}}}}%
    \put(0.83727645,0.26739803){\color[rgb]{0,0,0}\makebox(0,0)[lt]{\lineheight{0}\smash{\begin{tabular}[t]{l}$\mathcal{Y}_2$\end{tabular}}}}%
    \put(0.71144081,0.10493872){\color[rgb]{0,0,0}\makebox(0,0)[lt]{\lineheight{0}\smash{\begin{tabular}[t]{l}$\mathcal{Y}_3$\end{tabular}}}}%
    \put(0.609082,0.14625788){\color[rgb]{0,0,0}\makebox(0,0)[lt]{\lineheight{0}\smash{\begin{tabular}[t]{l}$\mathcal{Y}_4$\end{tabular}}}}%
    \put(0.44361208,0.30010221){\color[rgb]{0,0,0}\makebox(0,0)[lt]{\lineheight{1.25}\smash{\begin{tabular}[t]{l}{\Large Flow $f$}\end{tabular}}}}%
  \end{picture}%
\endgroup%

%% file: _background.tex

\textbf{Normalizing flows} \citep{rezende2015} define a probability density $p(\vy)$ using an invertible transformation $f$ between $\vy$ and a latent $\vz$ with a base distribution $p(\vz)$, i.e. 
\begin{equation*}
    \vy = f^{-1}(\vz) \quad\text{where}\quad \vz \sim p(\vz),
\end{equation*}
The density of $\vy$ can be computed as
\begin{equation*}
    p(\vy) = p(\vz) \left| \det \frac{\partial \vz}{\partial \vy} \right| = p(f(\vy)) \left| \det \frac{\partial f(\vy)}{\partial \vy} \right|.
\end{equation*}
The main challenge in designing flows is to develop transformations $f$ that are flexible, yet invertible and with cheap-to-compute Jacobian determinants.
Luckily, more expressive flows can be obtained through a composition $f = f_K \circ ... \circ f_2 \circ f_1$ of simpler flow layers $f_1, f_2, ..., f_K$. The computation cost of the forward pass, the inverse pass and the Jacobian determinant for the composition will simply be the sum of costs for the components. 
While this compositional approach to building expressive densities make flow models attractive, they are not directly applicable to discrete data. Consequently, a method known as \emph{dequantization} is typically employed.

\textbf{Uniform dequantization} refers to the process of converting discrete $\vx \in \{0,1,2,...,255\}^D$ to a continuous $\vy \in [0,256]^D$ by simply adding uniform noise, i.e.
\begin{equation*}
    \vy = \vx + \vu \quad\text{where}\quad \vu \sim \prod_{d=1}^D \mathrm{Unif}\left(u_d|0,1\right).
\end{equation*}
This ensures that the values fill the continuous space $[0,256]^D$ and consequently that continuous models will not collapse towards point masses at the discrete points during training. Uniform dequantization was proposed by \citet{uria2013} with exactly this motivation.
\citet{theis2016} further showed that optimizing a continuous model on uniformly dequantized samples corresponds to maximizing a lower bound on a discrete log-likelihood.

\textbf{Variational dequantization} was introduced by \citet{ho2019} as a generalization of uniform dequantization based on variational inference. Let $p(\vy)$ be some flexible continuous model and assume an observation model of the form $P(\vx | \vy) = \I(\vy \in \mathcal{B}(\vx))$,
where $\mathcal{B}(\vx)$ is the region in $\mathcal{Y}$ associated with $\vx$, e.g. a hypercube with one corner in $\vx$, i.e. $\{\vx + \vu | \vu \in [0,1)^D\}$.

As shown by \citet{ho2019}, using a dequantization distribution $q(\vy|\vx)$, one can develop a lower bound on the discrete log-likelihood $\log P(\vx)$ using Jensen's inequality,
\begin{align*}
    \log P(\vx) &= \log \int P(\vx | \vy) p(\vy) d\vy = \log \int_{\vy \in \mathcal{B}(\vx)} p(\vy) d\vy \\
    &= \log \int_{\vy \in \mathcal{B}(\vx)} q(\vy | \vx) \frac{p(\vy)}{q(\vy | \vx)} d\vy \geq \int_{\vy \in \mathcal{B}(\vx)} q(\vy | \vx) \log \frac{p(\vy)}{q(\vy | \vx)} d\vy.
\end{align*}
This corresponds exactly to the \emph{evidence lower bound} (ELBO) used in variational inference, where the dequantization distribution $q(\vy|\vx)$ coincides with the usual variational posterior approximation. 

Note that for uniform dequantization,
$q(\vy | \vx) = \prod_{d=1}^D \mathrm{Unif}(y_d | x_d, x_d+1)$,
the bound simplifies to
$\log P(\vx) \geq \E_{q(\vy | \vx)} [\log p(\vy)]$ since $q(\vy | \vx)=1$ over the entire integration region $\mathcal{B}(\vx)$.
Training with this lower bound corresponds to the common procedure for training flows on discrete data, i.e. fit the continuous density $p(\vy)$ to uniformly dequantized samples $\vy$. \citet{ho2019} proposed to use a more flexible flow-based dequantization distribution $q(\vy|\vx)$ in order to tighten the bound. The bound can further be tightened by using the importance weighted bound (IWBO) of \citet{burda2016}. In Sec. \ref{sec:dequantization_gap}, we identify a class of flows which allow direct computation of $\log P(\vx)$ instead of a lower bound.

%% file: _subset_flows.tex

In this section, we define the \emph{dequantization gap}, the difference between the discrete log-likelihood and its variational lower bound due to dequantization. Next, we introduce \emph{subset flows}, a class of flows for which dequantization is not needed, allowing us to directly optimize the discrete likelihood.

\subsection{The Dequantization Gap}

Flow models such as RealNVP \citep{dinh2017} and Glow \citep{kingma2018} have achieved remarkable performance for image data while still allowing efficient sampling with impressive sample quality.
However, in terms of log-likelihood, they still lag behind autoregressive models \citep{ho2019, ma2019}. While some of the performance gap might be the result of less expressive models, much of the gap seems to stem from a loose variational bound, as demonstrated by \citet{ho2019} and \citet{ma2019}. We term the difference between the discrete log-likelihood and its lower bound the \emph{dequantization gap}:
\begin{align*}
    \mathrm{Deq.\,Gap:} = \log P(\vx) - \E_{q(\vy|\vx)}\left[\log \frac{p(\vy)}{q(\vy|\vx)}\right] = \mathbb{D}_{KL} \left[q(\vy|\vx) \| p(\vy|\vx)\right].
\end{align*}

In the next subsection, we will introduce \emph{subset flows} which allow the discrete likelihood to be computed in closed form. This completely closes the dequantization gap and allows us to recover existing autoregressive models as flow models.

\subsection{Subset Flows}
\label{subsec:subset1d}

Dequantization facilitates computation of a lower bound of the discrete log-likelihood $\log P(\vx)$. However, using conservation of probability measure, we may compute the exact likelihood as
\begin{align*}
    P(\vx) = \int P(\vx | \vy) p(\vy) d\vy
    &= \int_{\vy \in \mathcal{B}(\vx)} p(\vy) d\vy = \int_{\vz \in f(\mathcal{B}(\vx))} p(\vz) d\vz,
\end{align*}
where $f(\mathcal{B}(\vx))$ is the image of $f$ applied to $\mathcal{B}(\vx)$, i.e. $f(\mathcal{B}(\vx)) = \{f(\vy) | \vy \in \mathcal{B}(\vx) \}$.
Assuming a standard uniform base distribution, $p(\vz) = \prod_{d=1}^D \mathrm{Unif}(z_d | 0,1)$, this formula takes the simple form
\begin{equation}
    \label{eq:volume}
     P(\vx) = \int_{\vz \in f(\mathcal{B}(\vx))} d\vz = \mathrm{Volume}(f(\mathcal{B}(\vx))).
\end{equation}
Interestingly, in order to compute $\log P(\vx)$, we do not need to keep track of infinitesimal volume changes with a Jacobian determinant. Instead, we have to keep track of the finite volume changes of the set $\mathcal{B}(\vx)$. While Eq. \ref{eq:volume} applies to any flow $f$ in principle, the computation is generally intractable. We define \emph{subset flows} as the class of flows $f: \mathcal{Y} \rightarrow \mathcal{Z}$ which have the additional property that they can tractably transform subsets of the input space, $\mathcal{Y}_s \subset \mathcal{Y}$, to subsets in the latent space, $\mathcal{Z}_s \subset \mathcal{Z}$, 
This is illustrated in Figure \ref{fig:subset_flows}. By keeping track of how the finite volume $\mathcal{B}(\vx)$ is transformed to a finite volume $f(\mathcal{B}(\vx))$ in the latent space, subset flows facilitate exact computation of the discrete likelhood in Eq. \ref{eq:volume}.

\textbf{Subset Flows in 1D.} In 1D, the computation in  Eq. \ref{eq:volume} with uniform $p(z)$ is particularly simple: 
\begin{align*}
    P(x) &= \int P(x | y) p(y) dy = \int_{y=x}^{x+1} p(y) dy = \int_{z=f(x)}^{f(x+1)} p(z) dz = f(x+1) - f(x),
\end{align*}
where $f$ then must correspond to the \emph{cumulative distribution function} (CDF) of $p(y)$.

\textbf{Autoregressive Subset Flows.} Subset flows present a different set of challenges compared to regular flows. In order to compute the discrete likelihood, we need not worry about computation of Jacobian determinants. Instead, we need flows where we can keep track of a finite volume. One straightforward approach to do this in higher dimensions is to work solely with hyperrectangles. Hyperrectangles have the benefit that they can easily be represented using two extreme points of the hyperrectangle. Furthermore, we can efficiently compute the volume of a hyperrectangle.

In order to work entirely with hyperrectangles, we need: 1) to partition the continuous space $\mathcal{Y}$ into hyperrectangles $\mathcal{B}(\vx)$ and 2) a flow $f$ such that the regions $f(\mathcal{B}(\vx))$ resulting from $f$ remain hyperrectangles. 
For the first point with e.g. $\mathcal{Y} = [0,256]^D$, we can define $\mathcal{B}(\vx) = \{\vx + \vu | \vu \in [0,1)^D\}$, resulting in disjoint hypercubes for each of the discrete values. The second point can be achieved by using an autoregressive flow with what we denote \emph{bin conditioning}. 

\begin{wrapfigure}{r}{0.5\linewidth}
    \centering
    \vspace{-5pt}
    \setlength{\belowcaptionskip}{-35pt}
    \resizebox{1.0\linewidth}{!}{
    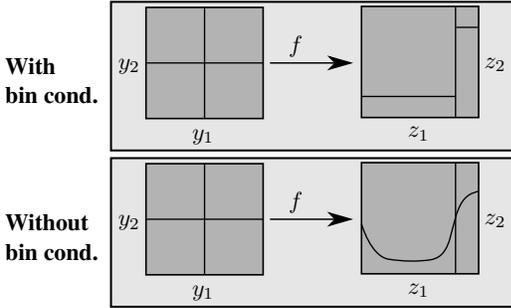
    }
    \caption{The effect of bin conditioning for a 2-dimensional binary problem. For the transformation with bin conditioning, the transformed rectangles remain rectangles.}
    \label{fig:bin_cond}
\end{wrapfigure}

\textbf{Bin conditioning} is achieved by conditioning on \emph{the bin to which a value belongs rather than its exact value}. For the transformation of dimension $d$, this is achieved by
\begin{align*}
    z_d^{\mathrm{(lower)}} &= f\left(y_d^{\mathrm{(lower)}} | \vlambda_d\left(\vy_{1:d-1}^{\mathrm{(lower)}}\right) \right), \\
    z_d^{\mathrm{(upper)}} &= f\left(y_d^{\mathrm{(upper)}} | \vlambda_d\left(\vy_{1:d-1}^{\mathrm{(lower)}}\right) \right),
\end{align*}
where $[y_d^{\mathrm{(lower)}}, y_d^{\mathrm{(upper)}}]$ are the boundaries of the input hyperrectangle and $[z_d^{\mathrm{(lower)}}, z_d^{\mathrm{(upper)}}]$ the output hyperrectangle. Importantly, the parameters $\vlambda_d$ are conditioned on the lower corner of the bin, $\vy_{1:d-1}^{\mathrm{(lower)}}$, rather than the exact value $\vy_{1:d-1}$, thus resulting in the same parameters $\vlambda_d$ regardless of the exact value of $\vy_{1:d-1}$ within the bin. This is an instance of bin conditioning and ensures that the output region $f(\mathcal{B}(\vx))$ will remain a hyperrectangle. Fig. \ref{fig:bin_cond} illustrates the effect of bin conditioning in a 2-dimensional binary problem. Note that conditioning on the upper corner or on both corners also constitute valid choices.

%% file: figures/illustrations/bin_cond.pdf_tex
\begingroup%
  \makeatletter%
  \providecommand\color[2][]{%
    \errmessage{(Inkscape) Color is used for the text in Inkscape, but the package 'color.sty' is not loaded}%
    \renewcommand\color[2][]{}%
  }%
  \providecommand\transparent[1]{%
    \errmessage{(Inkscape) Transparency is used (non-zero) for the text in Inkscape, but the package 'transparent.sty' is not loaded}%
    \renewcommand\transparent[1]{}%
  }%
  \providecommand\rotatebox[2]{#2}%
  \newcommand*\fsize{\dimexpr\f@size pt\relax}%
  \newcommand*\lineheight[1]{\fontsize{\fsize}{#1\fsize}\selectfont}%
  \ifx\svgwidth\undefined%
    \setlength{\unitlength}{216.8503937bp}%
    \ifx\svgscale\undefined%
      \relax%
    \else%
      \setlength{\unitlength}{\unitlength * \real{\svgscale}}%
    \fi%
  \else%
    \setlength{\unitlength}{\svgwidth}%
  \fi%
  \global\let\svgwidth\undefined%
  \global\let\svgscale\undefined%
  \makeatother%
  \begin{picture}(1,0.5979358)%
    \lineheight{1}%
    \setlength\tabcolsep{0pt}%
    \put(0,0){\includegraphics[width=\unitlength,page=1]{bin_cond.pdf}}%
    \put(-0.0014699,0.15109859){\color[rgb]{0,0,0}\makebox(0,0)[lt]{\lineheight{1.25}\smash{\begin{tabular}[t]{l}\textbf{Without} \\\textbf{bin cond.}\end{tabular}}}}%
    \put(0,0){\includegraphics[width=\unitlength,page=2]{bin_cond.pdf}}%
    \put(0.55097585,0.19331826){\color[rgb]{0,0,0}\makebox(0,0)[lt]{\lineheight{1.25}\smash{\begin{tabular}[t]{l}$f$\end{tabular}}}}%
    \put(0,0){\includegraphics[width=\unitlength,page=3]{bin_cond.pdf}}%
    \put(0.36250433,0.02382343){\color[rgb]{0,0,0}\makebox(0,0)[lt]{\lineheight{1.25}\smash{\begin{tabular}[t]{l}$y_1$\end{tabular}}}}%
    \put(0.21849507,0.16083679){\color[rgb]{0,0,0}\makebox(0,0)[lt]{\lineheight{1.25}\smash{\begin{tabular}[t]{l}$y_2$\end{tabular}}}}%
    \put(0.78224758,0.02803725){\color[rgb]{0,0,0}\makebox(0,0)[lt]{\lineheight{1.25}\smash{\begin{tabular}[t]{l}$z_1$\end{tabular}}}}%
    \put(0.93169665,0.16281325){\color[rgb]{0,0,0}\makebox(0,0)[lt]{\lineheight{1.25}\smash{\begin{tabular}[t]{l}$z_2$\end{tabular}}}}%
    \put(0,0){\includegraphics[width=\unitlength,page=4]{bin_cond.pdf}}%
    \put(-0.0014699,0.45557935){\color[rgb]{0,0,0}\makebox(0,0)[lt]{\lineheight{1.25}\smash{\begin{tabular}[t]{l}\textbf{With} \\\textbf{bin cond.}\end{tabular}}}}%
    \put(0,0){\includegraphics[width=\unitlength,page=5]{bin_cond.pdf}}%
    \put(0.5509758,0.49779912){\color[rgb]{0,0,0}\makebox(0,0)[lt]{\lineheight{1.25}\smash{\begin{tabular}[t]{l}$f$\end{tabular}}}}%
    \put(0,0){\includegraphics[width=\unitlength,page=6]{bin_cond.pdf}}%
    \put(0.36250433,0.32830419){\color[rgb]{0,0,0}\makebox(0,0)[lt]{\lineheight{1.25}\smash{\begin{tabular}[t]{l}$y_1$\end{tabular}}}}%
    \put(0.21849507,0.46531765){\color[rgb]{0,0,0}\makebox(0,0)[lt]{\lineheight{1.25}\smash{\begin{tabular}[t]{l}$y_2$\end{tabular}}}}%
    \put(0.78224758,0.33251811){\color[rgb]{0,0,0}\makebox(0,0)[lt]{\lineheight{1.25}\smash{\begin{tabular}[t]{l}$z_1$\end{tabular}}}}%
    \put(0.93169675,0.46729401){\color[rgb]{0,0,0}\makebox(0,0)[lt]{\lineheight{1.25}\smash{\begin{tabular}[t]{l}$z_2$\end{tabular}}}}%
    \put(0,0){\includegraphics[width=\unitlength,page=7]{bin_cond.pdf}}%
  \end{picture}%
\endgroup%

%% file: _pixelcnn_as_flow.tex

In this section, we will show that several existing discrete autoregressive models, including WaveNet, PixelCNN and PixelCNN++, can be obtained as single-layer autoregressive flows, giving them a notion of a latent space and enabling their use as layers in a multi-layer flow. 

\textbf{Autoregressive models} excel at modeling discrete data $\vx \in \{0,1,...,255\}^D$ such as images, audio and video since they can directly model discrete distributions.
Numerous models of this form have been proposed in recent years \citep{oord2016pixelrnn, oord2016gatedpixelcnn, oord2016wavenet, kalchbrenner2017, salimans2017, parmar2018, chen2018, menick2019, child2019}. These models rely on autoregressive neural networks constructed using masked convolutions and/or masked self-attention layers and have constituted the state-of-the-art in terms of log-likelihood. 

\newcommand{\sod}{.32}
\newcommand{\plod}{t}

\begin{figure}[t]
    \centering
    \begin{subfigure}[\plod]{\sod\linewidth}
    \tcbox[sharp corners,boxsep=0.0mm, boxrule=0.2mm,
           top=0pt,left=0pt,right=0pt,bottom=0pt,
           colframe=black, colback=white]
           {\includegraphics[width=\textwidth]{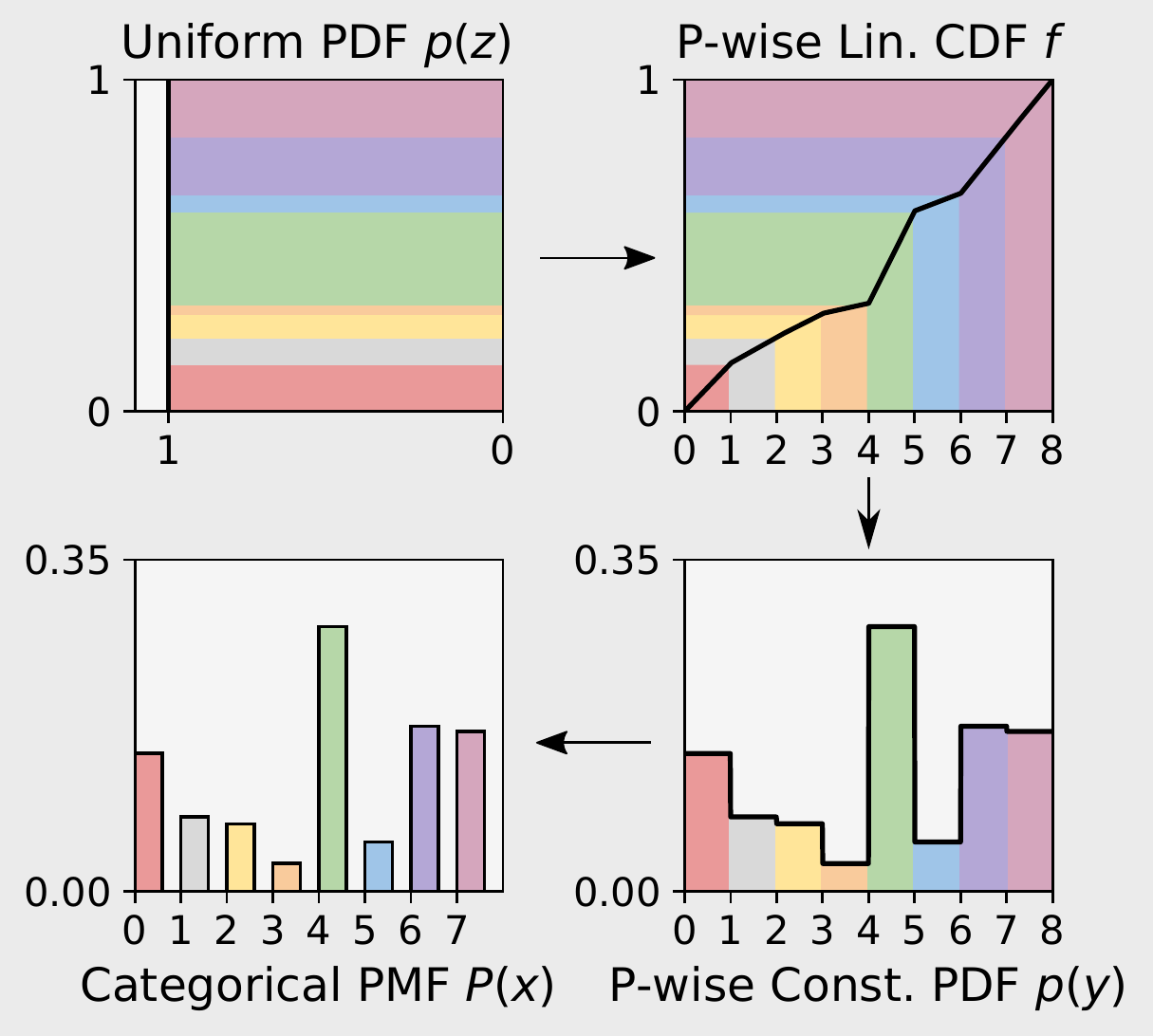}}
    \caption{Categorical}
    \label{fig:subset_flows1d_cat}
    \end{subfigure}
    \begin{subfigure}[\plod]{\sod\linewidth}
    \tcbox[sharp corners,boxsep=0.0mm, boxrule=0.2mm,
           top=0pt,left=0pt,right=0pt,bottom=0pt,
           colframe=black, colback=white]
           {\includegraphics[width=\textwidth]{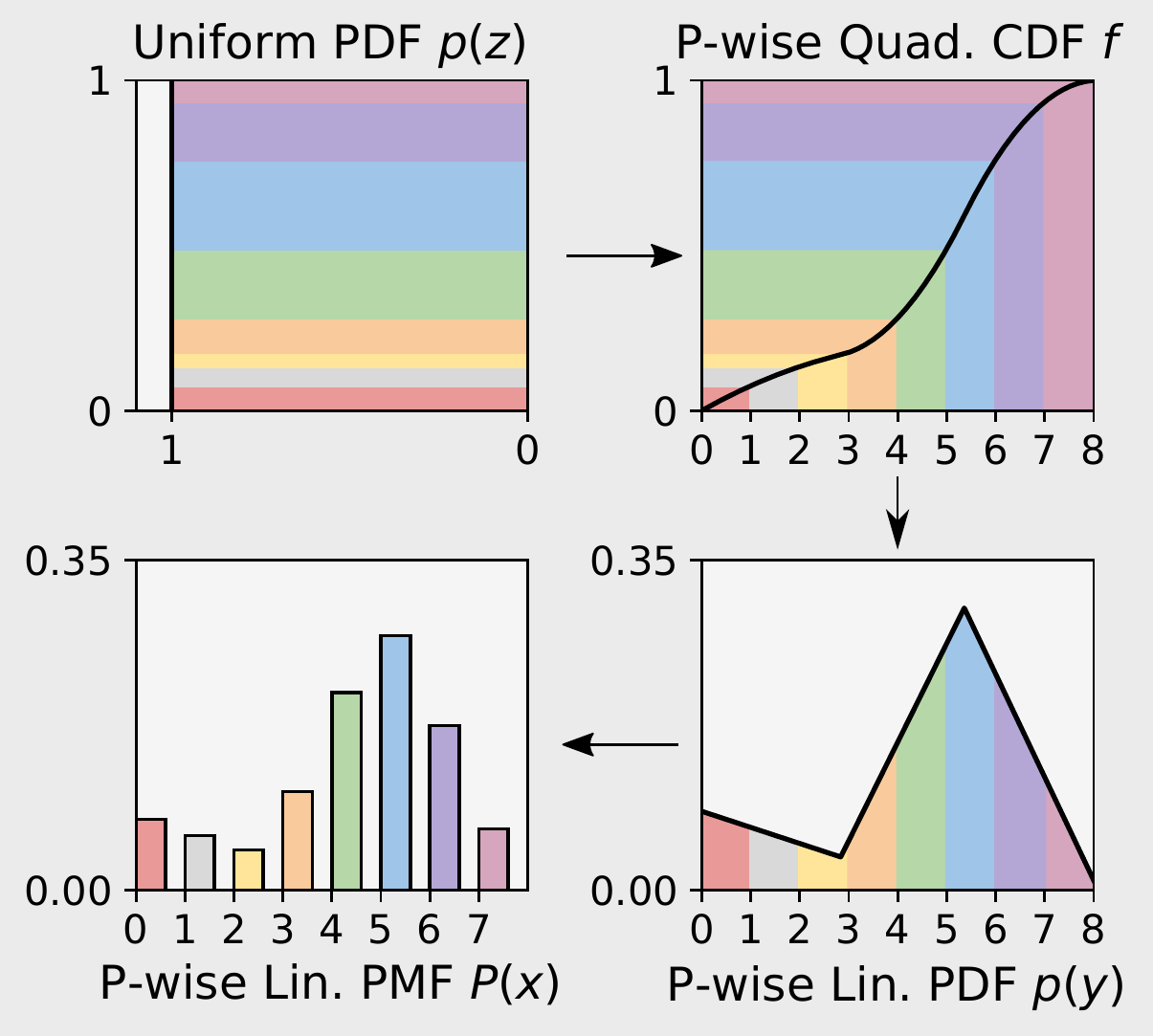}}
    \caption{Discretized Piecewise Linear}
    \label{fig:subset_flows1d_quad}
    \end{subfigure}
    \begin{subfigure}[\plod]{\sod\linewidth}
    \tcbox[sharp corners,boxsep=0.0mm, boxrule=0.2mm,
           top=0pt,left=0pt,right=0pt,bottom=0pt,
           colframe=black, colback=white]
           {\includegraphics[width=\textwidth]{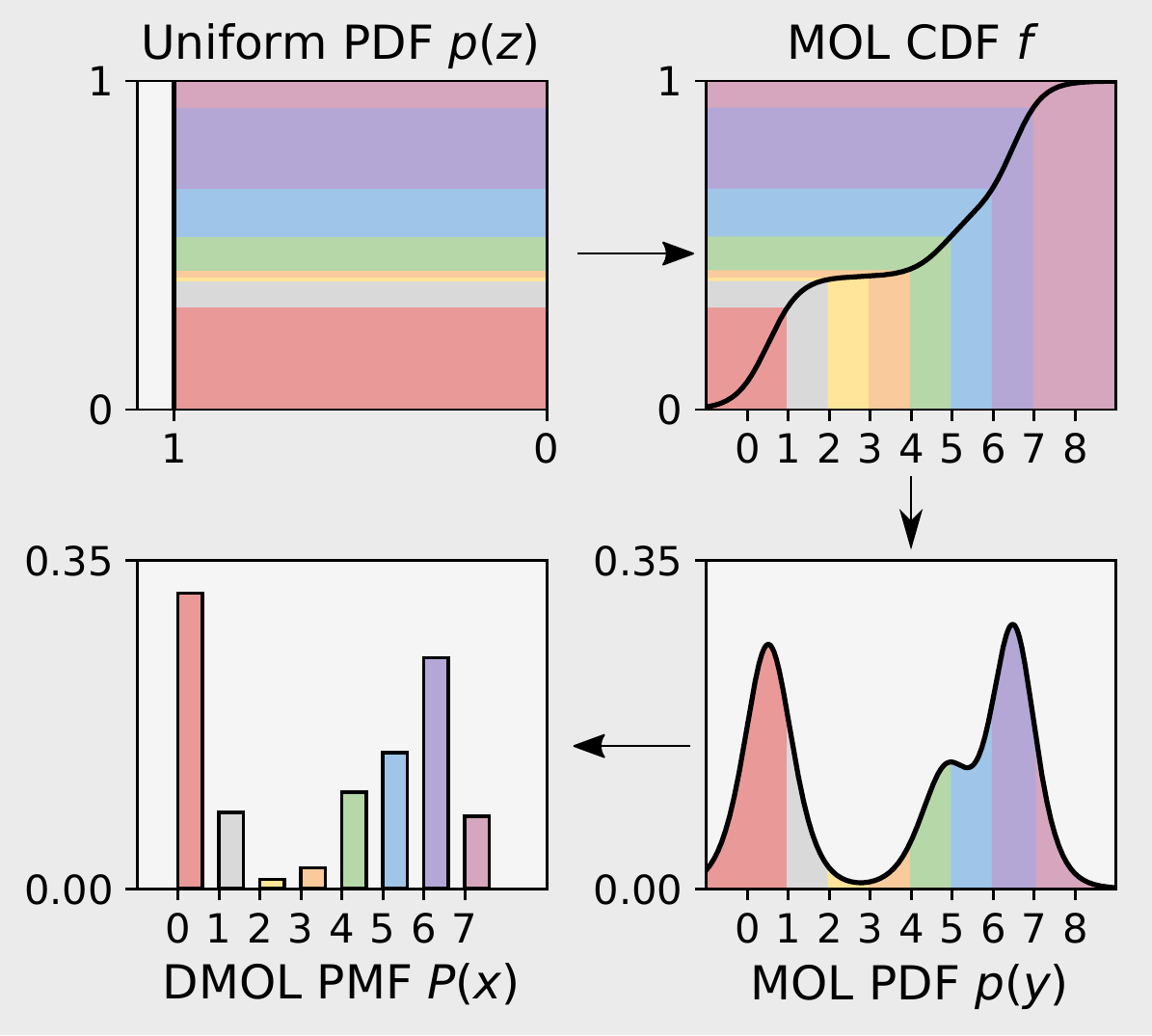}}
    \caption{Discretized Mixture of Logistics}
    \label{fig:subset_flows1d_dmol}
    \end{subfigure}
    
    \caption{Categorical, Discretized Piecewise Linear and Discretized Mixture of Logistics distributions as 1D subset flows. The \emph{arrows} indicate the direction for \textbf{generating samples}: \emph{1)} sample uniform noise $z$, \emph{2)} pass $z$ through the inverse flow/CDF $f^{-1}$ to obtain a continuous sample $y$, \emph{3)} quantize $y$ to obtain a discrete sample $x$. For subset flows, we can tractably invert this process to compute likelihoods. The \emph{colors} illustrate the flow of mass when \textbf{computing the likelihood}: \emph{1)} determine the region $\mathcal{B}(x)$ associated with observation $x$, \emph{2)} pass the region through the flow (in 1D, pass the two extremes of the region through), \emph{3)} compute the volume of the latent region. Note that while subset flows are straightforward in 1D, some care must be taken to extend them to higher dimensions.}
    \label{fig:subset_flows1d}
\end{figure}

\textbf{PixelCNN} and related models \citep{oord2016pixelrnn, oord2016gatedpixelcnn, oord2016wavenet, kalchbrenner2017, menick2019, child2019} take a simple approach to modelling ordinal discrete data: they use autoregressive networks to parameterize Categorical distributions, i.e.
$P(\vx) = \prod_{d=1}^D \mathrm{Cat}\left(x_d | \vx_{1:d-1}\right)$.
The Categorical distribution with $K$ categories may be obtained using subset flows as follows: Define a uniform base distribution, let $\mathcal{Y}=[0,K)$ and specify a piecewise linear CDF $f(y)$ \citep{muller2018}, 
\begin{equation*}
    f(y) = (y-(k-1)) \pi_k +  \sum_{l=1}^{k-1} \pi_l, \qquad\mathrm{for}\,\, k-1 \leq y < k
\end{equation*}
where $\pi_1,...,\pi_K \geq 0$, $\sum_{k=1}^K \pi_k = 1$. 
This yields a piecewise constant density $p(y)$, which upon quantization yields the Categorical distribution (see Fig. \ref{fig:subset_flows1d_cat}). Using Eq. \ref{eq:volume}, we find the discrete likelihood to be $P(x) = f(x+1) - f(x) = \pi_k$. 
See App. \ref{app:splines} for more details.
PixelCNN may thus be obtained as an autoregressive flow by using 1) a uniform base distribution, 2) bin conditioning, and 3) piecewise linear transformations, also known as linear splines, as the elementwise transformations. The result is an autoregressive subset flow which corresponds \emph{exactly} to the original PixelCNN model. 

\textbf{Higher order splines} such as quadratic, cubic or rational-quadratic \citep{muller2018, durkan2019} can be used as replacement of the linear, yielding novel models. The distribution obtained from quadratic splines is illustrated in Fig. \ref{fig:subset_flows1d_quad} (see App. \ref{app:splines} for more details). In our experiments, we show that quadratic splines tend to improve performance over linear splines.

\textbf{PixelCNN++} and related models \citep{salimans2017, parmar2018, chen2018} make use the \emph{Discretized Mixture of Logistics} (DMOL) \citep{salimans2017} distribution,
$P(\vx) = \prod_{d=1}^D \mathrm{DMOL}\left(x_d | \vx_{1:d-1}\right)$,
The DMOL distribution can be obtained using subset flows as follows: Define a uniform base distribution and let $f$ be the CDF of a mixture of logistics distribution, i.e.
\begin{equation*}
    f(y) = \sum_{m=1}^M \pi_m \sigma\left(\frac{y - 0.5 - \mu_m}{s_m}\right).
\end{equation*}
With bin boundaries defined at $y \in \{-\infty,1,2,...,255,\infty\}$, the discrete likelihood is
\footnotesize 
\begin{align*}
    &P(x) = f(y^{\mathrm{(upper)}}) - f(y^{\mathrm{(lower)}}) =
    \begin{cases}
        \sum\limits_{m=1}^M \!\pi_m\! \left[ \sigma\!\left(\frac{0.5 - \mu_m}{s_m}\right) \right]\!, x\!=\!0 \\
        \sum\limits_{m=1}^M \!\pi_m\! \left[ \sigma\!\left(\frac{x + 0.5 - \mu_m}{s_m}\right) \!-\! \sigma\!\left(\frac{x - 0.5 - \mu_m}{s_m}\right) \right]\!, x\!=\!1\!,\!...,\!254 \\
        \sum\limits_{m=1}^M \!\pi_m\! \left[ 1\! - \!\sigma\!\left(\frac{255 - 0.5 - \mu_m}{s_m}\right) \right]\!, x\!=\!255
    \end{cases}
\end{align*}
\normalsize
corresponding exactly to the DMOL as defined in \citep{salimans2017} (illustrated in Fig. \ref{fig:subset_flows1d_dmol}).

In practice, PixelCNN++ makes use of a multivariate version of the DMOL distribution. For an image with $D=CS$ dimensions, $C$ channels and $S$ spatial locations, the $C$-dimensional distribution for each of the $S$ spatial locations are modelled using a multivariate DMOL distribution. This multivariate DMOL distribution may itself be expressed as an autoregressive flow. See App. \ref{app:dmol} for more details. PixelCNN++ can thus be viewed as a \emph{nested autoregressive flow} where the network is autoregressive over the spatial dimensions and outputs parameters for the autoregressive flows along the channels. 

\textbf{Beyond single-layer flows.} By replacing the uniform base distribution by more flow layers, a more expressive distribution may be obtained. However, this will typically make the exact likelihood computation intractable, thus requiring dequantization. One exception is multi-layer autoregressive subset flows -- where the autoregressive order is the same for all layers -- which we consider in App.~\ref{app:multilayer_subset}. In our experiments, we show that compositions of PixelCNNs in flows yield powerful models.

%% file: _related_work.tex

This work is related to several lines of work. 
First of all, this work builds on work formulating autoregressive models as flows \citep{kingma2016, papamakarios2017, huang2018, oliva2018, jaini2019}. However, these only apply to continuous distributions and therefore do not include ordinal discrete autoregressive models such as the PixelCNN family of models \citep{oord2016pixelrnn, oord2016gatedpixelcnn, oord2016wavenet, kalchbrenner2017, salimans2017, parmar2018, chen2018, menick2019, child2019}. 

Second, this work builds on the variational view of dequantization \citep{theis2016, ho2019, hoogeboom2020}. \citet{uria2013} introduced uniform dequantization, \citet{theis2016} showed that this leads to a lower bound on the discrete likelihood and \citet{ho2019} further proposed to use a more flexible dequantization distribution in order to tighten the dequantization gap. We expand on this by showing that for subset flows, we can perform exact inference and thus close the dequantization gap completely.

Finally, one may model discrete data with flows that are discrete. \citet{hoogeboom2019} present discrete flows for ordinal integer data, while \citet{tran2019} present discrete flows for nominal categorical data. Both of these works make use of the straight-through estimator \citep{bengio2013} to backpropagate through the rounding operations, resulting in a gradient bias. Unlike these works, we make use of continuous flows, but apply them to discrete data. Consequently, we can compute exact gradients and therefore avoid the performance impacts arising from biased gradients.

%% file: _experiments.tex

\subsection{The Latent Space of PixelCNNs}
\input{_exp_interp.tex}

\subsection{The Effect of the Dequantization Gap}
\label{sec:exp_bound}
\input{_exp_bound.tex}

\subsection{PixelCNN in Flows}
\input{_exp_multi.tex}

%% file: _exp_interp.tex

\begin{wrapfigure}{r}{0.5\linewidth}
    \vspace{-15pt}
    \setlength{\belowcaptionskip}{-20pt}
    \centering
    \includegraphics[width=\linewidth]{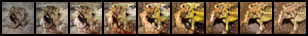}
    \includegraphics[width=\linewidth]{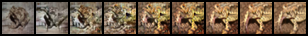}
    \includegraphics[width=\linewidth]{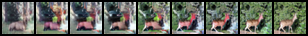}
    \includegraphics[width=\linewidth]{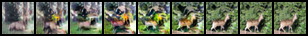}
    \includegraphics[width=\linewidth]{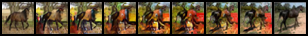}
    \includegraphics[width=\linewidth]{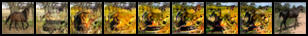}
    \includegraphics[width=\linewidth]{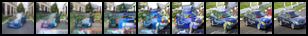}
    \includegraphics[width=\linewidth]{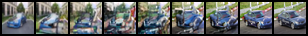}
    \caption{Latent space interpolations between pairs of CIFAR-10 test set images using PixelCNN (odd rows) and PixelCNN++ (even rows).
    These models are known for capturing local correlations well, but typically struggle with long-range dependencies. This is reflected in several of the interpolated images, which tend to lack global coherence.}
    \label{fig:interpolations_c10}
\end{wrapfigure}

PixelCNN, PixelCNN++ and related models are typically viewed as purely discrete autoregressive models which have no latent space associated with them. Our interpretation of these models as single-layer flows opens up for exploration of these existing models latent space. 
To illustrate this possibility, we trained PixelCNN \citep{oord2016pixelrnn} and PixelCNN++ \citep{salimans2017} as flow models on CIFAR-10, with results exactly matching the reported numbers of 3.14 and 2.92 bits/dim. 

Some examples of interpolations between CIFAR-10 test images are shown in Figure \ref{fig:interpolations_c10}. These were obtained by interpolating along a path of \emph{equally-probable} samples under the base distribution. See App. \ref{app:interp} for more details.

%% file: _exp_bound.tex

Our work shows that the PixelCNN family of models can be formulated as flow models where the dequantization gap between the true likelihood and the variational lower bound is completely closed. This suggests that they may be trained using either 1) uniform dequantization, 2) variational dequantization or 3) the exact likelihood. The resulting models should have decreasing dequantization gaps in the listed order.
Surveying results from the literature (collected in Table \ref{tab:existing} in App. \ref{app:collection_work}), we observe significant improvements between the categories with the best results for e.g. CIFAR-10 at $3.28$, $3.08$ and $2.80$, suggesting that the dequantization gap has a significant impact on results.


\begin{table*}[!ht]
\caption{The effect of the dequantization gap. We compare three models, PixelCNN, PixelCNN (Quad.) and PixelCNN++. For each model, we trained three versions, one using the exact likelihood and two using the ELBO with uniform dequantization, both with and without bin conditioning. The models trained using the ELBO are evaluated using 1) the ELBO, 2) the IWBO (importance weighted bound) \citep{burda2016}, and 3) the exact likelihood. See Sec. \ref{sec:exp_bound} for an explanation.}
\label{tab:bound}
\centering
\begin{tabular}{cccccc}
\hline
\textbf{Bin Cond.} & \textbf{Training} & \textbf{Eval.} & \textbf{PixelCNN} & \textbf{PixelCNN (Q)} & \textbf{PixelCNN++} \\
\hline
\multirow{4}{*}{No} & \multirow{4}{*}{ELBO} & ELBO & 3.248 & 3.251 & 3.112 \\
 & & IWBO(10) & 3.235 & 3.237 & 3.095 \\
 & & IWBO(100) & 3.227 & 3.228 & 3.086 \\
 & & IWBO(1000) & 3.221 & 3.223 & 3.079 \\
\hline
\multirow{5}{*}{Yes} & \multirow{5}{*}{ELBO} & ELBO & \textbf{3.141} & 3.142 & 2.993 \\
 & & IWBO(10) & \textbf{3.141} & 3.134 & 2.983 \\
 & & IWBO(100) & \textbf{3.141} & 3.129 & 2.978 \\
 & & IWBO(1000) & \textbf{3.141} & 3.126 & 2.974 \\
 & & Exact & \textbf{3.141} & 3.104 & 2.944 \\
\hline
Yes & Exact & Exact & \textbf{3.141} & \textbf{3.090} & \textbf{2.924} \\
\hline
\end{tabular}
\end{table*}

To further test this hypothesis, we make use of our flow interpretation of existing autoregressive models. We train three flow models on CIFAR-10: 1) PixelCNN, 2) PixelCNN with quadratic splines (Quad.) and 3) PixelCNN++ using three different setups:
\begin{enumerate}
    \item \emph{Exact likelihood:} We train models exploiting the fact that for subset flows we can compute exact likelihoods.
    \item \emph{Dequantization w/ bin cond.:} In this case, we train the exact same models as before, but we replace the exact likelihood objective with the ELBO. With this setup, we can investigate:
    \begin{itemize}
        \item The gap from the ELBO to the exact likelihood: $\log P(\vx | \vtheta_{\mathrm{ELBO}}) - \mathcal{L}(\vtheta_{\mathrm{ELBO}})$.
        \item How much closer the IWBO gets us in practice: $\log P(\vx | \vtheta_{\mathrm{ELBO}}) - \mathcal{L}_k(\vtheta_{\mathrm{ELBO}})$.
        \item The negative impact of training with the ELBO: $\log P(\vx | \vtheta_{\mathrm{Exact}}) - \log P(\vx | \vtheta_{\mathrm{ELBO}})$.
    \end{itemize}
    Here, $\vtheta$ denotes the model parameters, $\mathcal{L}(\vtheta)$ denotes the ELBO and $\mathcal{L}_k(\vtheta)$ the IWBO with $k$ importance samples for parameters $\vtheta$. Furthermore, $\vtheta_{\mathrm{ELBO}} = \argmax_{\vtheta} \mathcal{L}(\vtheta)$ and $\vtheta_{\mathrm{Exact}} = \argmax_{\vtheta} \log P(\vx | \vtheta)$.
    \item \emph{Dequantization w/o bin cond.:} We change the flows to not use bin conditioning. As a result, the latent regions will no longer be hyperrectangles and we therefore cannot compute exact likelihoods for these models. Note that this closely corresponds to how most flow models such as RealNVP and Glow are trained.
\end{enumerate}

The results are given in Table \ref{tab:bound}.
Some things to note from these results are:
\begin{itemize}
    \setlength\itemsep{0.5mm}
    \item The exact models match the reported numbers in \citet{oord2016pixelrnn} and \citet{salimans2017} at 3.14 and 2.92 bits/dim.
    \item Training with the ELBO negatively impacts performance, even when evaluating using the exact likelihood. Gaps of 0.014 and 0.020 bits/dim are found for PixelCNN (Quad.) and PixelCNN++.
    \item For models with bin conditioning trained with the ELBO, we can here compute the exact dequantization gap. For PixelCNN (Quad.) and PixelCNN++, this gap is found to be 0.038 and 0.049 bits/dim.
    \item The IWBO improves the estimate of $\log P(\vx)$ with an increasing number of importance samples. However, even for 1000 samples, less than half the gap has been closed, with 0.022 and 0.030 bits/dim remaining.
    \item For PixelCNN with bin conditioning, training with the ELBO does not impact performance. Here, the exact $p(\vy|\vx)$ is uniform and therefore exactly matches the uniform dequantization distribution $q(\vy|\vx)$, resulting in a dequantization gap of $\mathbb{D}_{KL} \left[q(\vy|\vx) \| p(\vy|\vx)\right]=0$.
    \item Models trained without bin conditioning show significantly worse performance with gaps of 0.107, 0.161 and 0.188 to the original exact models. This shows that the usual approach for training flows using uniform dequantization and no bin conditioning leads to significantly worse performance. 
\end{itemize}

%% file: _exp_multi.tex

\textbf{PixelFlow.} We now demonstrate that compositions of PixelCNNs in flows can yield expressive models. We first construct 2 models which we term \emph{PixelFlow} and \emph{PixelFlow++}. PixelFlow++ is a composition of a PixelCNN++ and a multi-scale flow architecture \citep{dinh2017} using 2 scales with 8 steps/scale. Each step is a composition of a coupling layer \citep{dinh2017} and an invertible 1$\times$1 convolution \citep{kingma2018}. Each coupling layer is parameterized by a DenseNet \citep{huang2017}.
PixelFlow uses the exact same setup as PixelFlow++, except it uses a quadratic spline version of PixelCNN instead of PixelCNN++. Both models make use of bin conditioning.

We train PixelFlow and PixelFlow++ using variational dequantization \citep{ho2019} and compare to other autoregressive and coupling flows trained with dequantization. The results are shown in Table \ref{tab:flows}. \emph{PixelFlow++ obtains state-of-the-art results for flow models trained with dequantization on CIFAR-10}. Samples from PixelFlow and PixelFlow++ are shown in Fig. \ref{fig:samples}. More samples can be found in App. \ref{app:samples}.

\begin{wrapfigure}{r}{0.39\linewidth}
\captionsetup{type=table}
\centering
\vspace{-4mm}
\caption{A flow of 4 quadratic spline PixelCNNs trained on CIFAR-10 with or without $90^{\circ}$ rotation using 1) uniform dequantization, 2) variational dequantization and 3) the exact likelihood.}
\label{tab:stack}
\begin{tabular}{lccc}
\toprule
\textbf{Rotation} & \textbf{Uni.} & \textbf{Var.} & \textbf{Exact} \\
\midrule
No & 3.066 & 3.026 & \textbf{3.012} \\
Yes  & 3.058 & \textbf{3.012} & - \\
\bottomrule
\end{tabular}
\end{wrapfigure}

\textbf{Stacks of PixelCNNs.} Next, we perform a series of experiments where we stack PixelCNNs in multi-layer flows with and without $90^{\circ}$ rotations in-between. Table \ref{tab:stack} shows results for stacks of 4 quadratic spline PixelCNNs. Note that when no rotation is used, the autoregressive order is the same for all the PixelCNNs. Consequently, we may use bin conditioning in all layers, yielding a multi-layer subset flow, which allows exact likelihood computation.
As expected, the models using rotation tend to perform better than those without.
Interestingly, however, the exact model \emph{without rotation} performs on par with the variational dequantization model \emph{with rotation}, which suffers from a non-zero dequantization gap. We further investigate multi-layer subset flows, which have dequantization gaps of exactly zero, in App. \ref{app:multilayer_subset}. Further details on all experiments can be found in App. \ref{app:exp_details}.

\newcommand{\fw}{0.48\linewidth}
\begin{figure}
\centering
\begin{minipage}[b]{0.38\textwidth}

\begin{subfigure}[b]{\fw}
    \centering
    \caption*{\textbf{PixelCNN}}
    \vspace{-2.8mm}
    \includegraphics[trim=0 136 136 0,clip,width=0.92\linewidth]{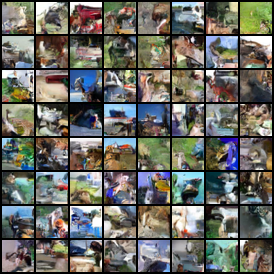}
\end{subfigure}
\begin{subfigure}[b]{\fw}
    \centering
    \caption*{\textbf{PixelCNN++}}
    \vspace{-2.8mm}
    \includegraphics[trim=0 136 136 0,clip,width=0.92\linewidth]{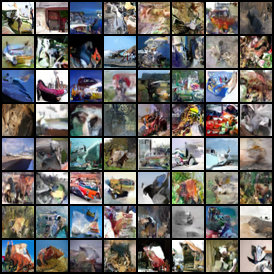}
\end{subfigure}

\begin{subfigure}[b]{\fw}
    \centering
    \caption*{\textbf{PixelFlow}}
    \vspace{-2.8mm}
    \includegraphics[trim=0 136 136 0,clip,width=0.92\linewidth]{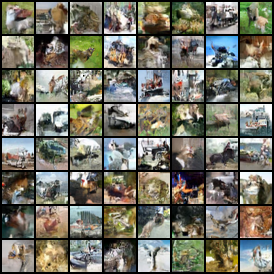}
\end{subfigure}
\begin{subfigure}[b]{\fw}
    \centering
    \caption*{\textbf{PixelFlow++}}
    \vspace{-2.8mm}
    \includegraphics[trim=0 136 136 0,clip,width=0.92\linewidth]{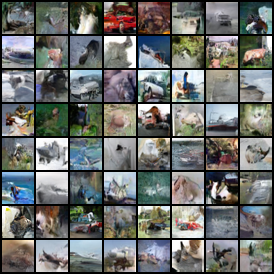}
\end{subfigure}

\caption{Unconditional samples.}\label{fig:samples}
\end{minipage}
\hspace{0.2cm}
\begin{minipage}[b]{0.58\textwidth}
\centering
\captionsetup{type=table}
\begin{tabular}{llc}
\toprule
\textbf{Model} & \textbf{AR} & $\leq $\textbf{Bits/dim} \\
\midrule
RealNVP \citep{dinh2017} & No & 3.49 \\
Glow \citep{kingma2018} & No & 3.35 \\
Flow++ \citep{ho2019} & No & 3.08 \\
\midrule
MintNet \citep{song2019} & Yes & 3.32 \\
MaCow \citep{cao2019} & Yes & 3.16 \\
\midrule
PixelFlow (ours) & Yes & 3.04 \\
PixelFlow++ (ours) & Yes & \textbf{2.92} \\
\bottomrule
\end{tabular}
\caption{Flow models trained with dequantization on CIFAR-10. PixelFlow(++) correspond to a composition of PixelCNN(++) and a Glow-like coupling flow.}\label{tab:flows}
\end{minipage}
\end{figure}

%% file: _conclusion.tex

We presented subset flows, a class of flows which can tractably transform finite volumes, a property that allow their use for ordinal discrete data like images and audio without the need for dequantization.
Based on subset flows, we could explicitly formulate existing autoregressive models such as PixelCNNs and WaveNets as single-layer autoregressive flows. Using this formulation of PixelCNNs, we were able to quantify exactly the performance impacts of training and evaluating flow models using dequantization. We further demonstrated that expressive flow models can be obtained using PixelCNNs as layers in multi-layer flows.

Potential directions for future work include designing novel forms of subset flows and developing novel state-of-the-art flow architectures using the formulation of the PixelCNN family as flows.

%% file: _acknowledgements.tex

We thank Emiel Hoogeboom, Priyank Jaini, Shihan Wang and Sindy L\"{o}we for helpful feedback.

%% file: _funding_disclosure.tex
This research was supported by the NVIDIA Corporation with the donation of TITAN X GPUs.

%% file: _broader_impact.tex

This is foundational research in generative models/unsupervised learning with proposal for new flow models and interpretation of existing autoregressive models as flow models. These models can be applied to for example unsupervised learning on images and audio. 
Unsupervised learning has the potential to greatly reduce the need for labeled data and thus improve models in applications such as medical imaging where a lack of data can be a limitation. 
However, it may also potentially be used to improve deep fakes with potentially malicious applications.

%% file: _appendix.tex


\section{Linear and Quadratic Splines as Flows}
\label{app:splines}

For self-containedness, we will here summarize linear and quadratic spline flows, i.e. piecewise linear and quadratic flows, as presented by \citet{muller2018}. 

\subsection{Linear Spline Flows}

Consider a univariate flow $f:\mathcal{Y}\rightarrow \mathcal{Z}$ where $\mathcal{Y} = [0,Q]$ and the latent space $\mathcal{Z} = [0,1]$.
The transformation $f$ is piece-wise linear between the points $\{(y_k,z_k)\}_{k=0}^K$, where
\begin{align*}
    0 &\equiv y_0 < y_1 < y_2 < ... < y_K \equiv Q \\
    0 &\equiv z_0 < z_1 < z_2 < ... < z_K \equiv 1.
\end{align*}
For piece-wise linear flows, we fix $y_1, y_2, ..., y_{K-1}$ and parameterize $z_1, z_2, ..., z_{K-1}$ using probabilities $\pi_1, \pi_2, ..., \pi_{K-1}$,
\begin{equation*}
    z_k = \sum_{l=1}^k \pi_l ,
\end{equation*}
where $\sum_{l=1}^K \pi_l = 1$.

The forward, inverse and Jacobian determinant computation of a linear spline $f$ can be written as
\begin{align*}
    \mathrm{\textbf{Forward:}}&\quad z = f(y) = \sum_{k=1}^K \I\left(y_{k-1} \leq y < y_k\right) \left[ z_{k-1} + \pi_k \frac{y-y_{k-1}}{y_k-y_{k-1}}  \right] \\
    \mathrm{\textbf{Inverse:}}&\quad y = f^{-1}(z) = \sum_{k=1}^K \I\left(z_{k-1} \leq z < z_{k}\right) \left[ y_{k-1} + (y-y_{k-1})\frac{z-z_{k-1}}{\pi_k} \right] \\
    \mathrm{\textbf{Jac. det.:}}&\quad \left| \det \frac{df(y)}{dy} \right| = \prod_{k=1}^K \pi_k^{\I(y \in [y_{k-1}, y_k))}.
\end{align*}

\subsubsection{Relation to the Categorical Distribution}

To obtain the Categorical distribution, we 1) let $Q=K$, 2) fix $y_0=0, y_1=1, ..., y_K=K$ and 3) use a uniform base distribution,
\begin{align*}
    z &\sim \mathrm{Unif}(z|0,1) \\
    y &= f^{-1}(z | \vpi).
\end{align*}
This yields a piecewise constant density,
\begin{align*}
 p(y | \vpi) = \prod_{k=1}^K \pi_k^{\I(y \in [k-1, k))},
\end{align*}
which upon quantization yields the Categorical distribution,
\begin{align*}
 p(x | \vpi) = \mathrm{Cat}(x | \vpi) = \prod_{k=1}^K \pi_k^{\I(x = k)}.
\end{align*}

\subsection{Quadratic Spline Flows}

Consider again a univariate flow $f:\mathcal{Y}\rightarrow \mathcal{Z}$ where $\mathcal{Y} = [0,Q]$ and the latent space $\mathcal{Z} = [0,1]$.
The transformation $f$ is piece-wise quadratic between the points $\{(y_k,z_k)\}_{k=0}^K$, where
\begin{align*}
    0 &\equiv y_0 < y_1 < y_2 < ... < y_K \equiv Q \\
    0 &\equiv z_0 < z_1 < z_2 < ... < z_K \equiv 1.
\end{align*}
In the linear spline case, we fixed the locations $y_0, y_1, ..., y_K$. If this is not done, the objective becomes discontinuous and thus difficult to train with gradient-based optimizers.
In the quadratic case, however, we may let the bin locations $y_0, y_1, ..., y_K$ be free parameters to be learned \citep{muller2018}.

The parameters of the flow are given by vectors $\hat{\vw} \in \mathbb{R}^{K}$ and $\hat{\vv} \in \mathbb{R}^{K+1}$. 
The bin widths are computed as
\begin{equation*}
    \vw = Q \cdot \mathrm{softmax}(\hat{\vw}),
\end{equation*}
while the bin edges are computed as
\begin{equation*}
    \vv = \frac{\exp (\hat{\vv})}{\sum_{k=1}^K \frac{\exp (\hat{v}_k) + \exp (\hat{v}_{k+1})}{2} w_k}.
\end{equation*}
The bin locations are given by the sum of bin widths
\begin{equation*}
    y_k = \sum_{l=1}^k w_l.
\end{equation*}
For a quadratic spline, the density will be piece-wise linear. We can use this to compute the mass in bin $k$ between the lower extreme $y_{k-1}$ and some point $y^*$ can be computed as
\begin{align*}
    \int_{y=y_{k-1}}^{y^*} \left[ v_{k-1} + \frac{y - y_{k-1}}{y_k - y_{k-1}} (v_k - v_{k-1}) \right] dy
    &= \int_{t=0}^{\alpha}  \left[ v_{k-1} + t (v_k - v_{k-1}) \right] (y_k - y_{k-1}) dt \\
    &= \left[t v_{k-1} + \frac{1}{2} t^2 (v_k - v_{k-1}) \right]_{t=0}^{\alpha} (y_k - y_{k-1}) \\
    &= \left[ \alpha v_{k-1} + \frac{1}{2} \alpha^2 (v_k - v_{k-1}) \right] w_k.
\end{align*}
where $t \equiv \frac{y - y_{k-1}}{y_k - y_{k-1}}$ and $\alpha \equiv \frac{y^* - y_{k-1}}{y_k - y_{k-1}}$. The total mass in the bin can be found by setting $y^* = y_k$ or equivalently $\alpha = 1$ to obtain
\begin{equation*}
    \int_{y=y_{k-1}}^{y_k} \left[ v_{k-1} + \frac{y - y_{k-1}}{y_k - y_{k-1}} (v_k - v_{k-1}) \right] dy = \frac{v_{k-1} + v_k}{2}w_k
\end{equation*}
Using this, we find that
\begin{equation*}
    z_k = \sum_{l=1}^{k} \frac{v_{l-1} + v_l}{2}w_l
\end{equation*}

The forward, inverse and Jacobian determinant computation of the resulting quadratic spline flow $f$ can be written as
\begin{align*}
    \mathrm{\textbf{Forward:}}&\quad z = f(y) = \sum_{k=1}^K \I\left(y_{k-1} \leq y < y_k\right) \left[ z_{k-1} + w_k\left(\alpha_k v_{k-1} + \frac{1}{2} \alpha_k^2 (v_k - v_{k-1})\right) \right] \\
    \mathrm{\textbf{Inverse:}}&\quad y = f^{-1}(z) = \sum_{k=1}^K \I\left(z_{k-1} \leq z < z_k\right) \left[ y_{k-1} + w_k\frac{\sqrt{v_{k-1}^2 + 2\frac{v_k-v_{k-1}}{w_k}(z-z_{k-1})} - v_{k-1}}{v_k - v_{k-1}} \right] \\
    \mathrm{\textbf{Jac. det.:}}&\quad \left| \det \frac{df(y)}{dy} \right| = \left[ v_{k-1} + \alpha_k (v_k - v_{k-1}) \right]^{\I(y \in [y_{k-1}, y_k))},
\end{align*}
where 
\begin{equation*}
    \alpha_k = \frac{y - y_{k-1}}{y_k - y_{k-1}}.
\end{equation*}


\section{The Multivariate DMOL as a Flow}
\label{app:dmol}

We will show that the \emph{multivariate discretized mixture of logistics} (multivariate DMOL) distribution can be obtained as an autoregressive flow. 
First, we describe the distribution as it was presented in \citet{salimans2017}.

\subsection{The Multivariate DMOL Distribution}

The discretized logistic distribution can be written as
\begin{equation*}
    P(x) = \mathrm{DLogistic}(x|\mu,s)
    =
    \begin{cases}
        \sigma\left(\frac{0.5 - \mu}{s}\right), x=0. \\
        \sigma\left(\frac{x + 0.5 - \mu}{s}\right) - \sigma\left(\frac{x - 0.5 - \mu}{s}\right), x=1,...,254. \\
        1 - \sigma\left(\frac{255 - 0.5 - \mu}{s}\right), x=255.
    \end{cases}
\end{equation*}
The univariate DMOL distribution uses this as mixture components,
\begin{equation*}
    P(x) = \mathrm{DMOL}(x|\vpi,\vmu,\vs) = \sum_{m=1}^M \pi_m \mathrm{DLogistic} (x|\mu_m, s_m).
\end{equation*}
The multivariate DMOL distribution, on the other hand, can be written as
\begin{equation*}
\begin{split}
    P(\vx) = \mathrm{MultiDMOL}(\vx|\vpi,\vmu,\vs,\vr) = \sum_{m=1}^M \pi_m &\mathrm{DLogistic} (x_3|\mu_{3,m}(x_1,x_2,\vr_m), s_{3,m}) \\
    &\mathrm{DLogistic} (x_2|\mu_{2,m}(x_1,\vr_m), s_{2,m}) \\
    &\mathrm{DLogistic} (x_1|\mu_{1,m}, s_{1,m}),
\end{split}
\end{equation*}
where
\begin{equation}
\label{eq:ar_mean}
\begin{split}
    \mu_{1,m} &= \mu_{1,m} \\
    \mu_{2,m}(x_1, \vr_m) &= \mu_{2,m} + r_{1,m} x_1 \\
    \mu_{3,m}(x_1, x_2, \vr_m) &= \mu_{3,m} + r_{2,m} x_1 + r_{3,m} x_2.
\end{split}
\end{equation}

\subsection{Rewriting the Multivariate DMOL Distribution}

We will now show how one can rewrite this distribution in an autoregressive form. Consider the usual 3-dimensional case and write the multivariate DMOL as
\begin{equation*}
    P(\vx) = \sum_{m=1}^M \pi_m P_m(x_3 | x_2, x_1) P_m(x_2 | x_1) P_m(x_1).
\end{equation*}
We can rewrite this distribution as
\begin{align*}
    P(\vx) &= \sum_{m=1}^M \pi_m P_m(x_3 | x_2, x_1) P_m(x_2 | x_1) P_m(x_1) \\
    &= \left[ \sum_{m=1}^M \pi_{3,m} P_m(x_3 | x_2, x_1) \right] \left[ \sum_{m=1}^M \pi_{2,m} P_m(x_2 | x_1) \right] \left[ \sum_{m=1}^M \pi_{1,m} P_m(x_1) \right] 
\label{eq:rewriting}
\end{align*}
where
\begin{equation}
\label{eq:ar_weights}
\begin{split}
    \pi_{1,m} &= \pi_m \\
    \pi_{2,m} &= \frac{\pi_m P_m(x_1)}{\sum_{m'=1}^M \pi_{m'} P_{m'}(x_1)} \\
    \pi_{3,m} &= \frac{\pi_m P_m(x_2 | x_1) P_m(x_1)}{\sum_{m'=1}^M \pi_{m'} P_{m'}(x_2 | x_1) P_{m'}(x_1)}.
\end{split}
\end{equation}

\subsection{The Multivariate DMOL Flow}

To summarize, we can write the multivariate DMOL as an autoregressive distribution with univariate DMOL conditionals,
\begin{equation*}
\begin{split}
    \mathrm{MultiDMOL}(\vx | \vpi, \vmu, \vs, \vr) = 
    &\,\,\,\, \mathrm{DMOL}(x_3|\vpi(x_2, x_1), \vmu(x_2, x_1, \vr), \vs) \\
    &\cdot \mathrm{DMOL}(x_2|\vpi(x_1), \vmu(x_1, \vr), \vs) \\
    &\cdot \mathrm{DMOL}(x_1|\vpi, \vmu, \vs),
\end{split}
\end{equation*}
where the means are given by Eq. \ref{eq:ar_mean} and the mixture weights by Eq. \ref{eq:ar_weights}.
Thus, we can obtain the multivariate DMOL flow as an autoregressive flow with the univariate DMOL flows from Sec. \ref{subsec:subset1d} as elementwise transformations.


\section{A Collection of Results from Previous Work}
\label{app:collection_work}

We here collect results from existing work in order to compare flow models trained with 1) uniform dequantization, 2) variational dequantization and 3) exact likelihood. The categories should have decreasing dequantization gaps in the listed order. The results are shown in Table \ref{tab:existing}. We observe significant improvements between the categories, suggesting that the dequantization gap has a significant impact on results.

\begin{table*}[!ht]
\caption{A collection of results from previous work (in bits/dim). We here divide the results into three categories, those trained with: 1) uniform dequantization, 2) variational dequantization and 3) exact likelihood. We observe that performance tends to drastically improve between the categories, suggesting the importance of the dequantization gap for flow models.}
\label{tab:existing}
\centering
\setlength{\tabcolsep}{3pt}
\begin{tabular}{ll@{\hspace{-2mm}}ccc}
\hline
\textbf{Training} & \textbf{Model} & \textbf{CIFAR-10} & \textbf{ImageNet32} & \textbf{ImageNet64} \\
\hline
\multirow{4}{*}{ELBO (U)} & RealNVP \citep{dinh2017} & 3.49 & 4.28 & 3.98 \\
 & Glow \citep{kingma2018} & 3.35 & \textbf{4.09} & 3.81 \\
 & MaCow \citep{ma2019} & \textbf{3.28} & - & \textbf{3.75} \\
 & Flow++ \citep{ho2019} & 3.29 & - & - \\
\hline
\multirow{2}{*}{ELBO (V)} & MaCow \citep{ma2019} & 3.16 & - & \textbf{3.69} \\
 & Flow++ \citep{ho2019} & \textbf{3.08} & \textbf{3.86} & \textbf{3.69} \\
\hline
\multirow{7}{*}{Exact} & PixelCNN \citep{oord2016pixelrnn} & 3.14 & - & - \\
 & Gated PixelCNN \citep{oord2016gatedpixelcnn} & 3.03 & 3.83 & 3.57 \\
 & PixelCNN++ \citep{salimans2017} & 2.92 & - & - \\
 & Image Transformer \citep{parmar2018} & 2.90 & \textbf{3.77} & - \\
 & PixelSNAIL \citep{chen2018} & 2.85 & 3.80 & 3.52 \\
 & SPN \citep{menick2019} & - & 3.79 & 3.52 \\
 & Sparse Transformer \citep{child2019} & \textbf{2.80} & - & \textbf{3.44} \\
\hline
\end{tabular}
\end{table*}


\section{Experimental Details}
\label{app:exp_details}

The code used for experiments is publicly available\footnote{\url{https://github.com/didriknielsen/pixelcnn_flow}}.
In our experiments, we used PixelCNN \citep{oord2016pixelrnn} and PixelCNN++ \citep{salimans2017} models. For hyperparameters, we followed the original architectures as closely as possible. The PixelCNN architecture we used for the CIFAR-10 experiments is summarized in Table \ref{tab:pcnn_arch}. 

\begin{table}[!ht]
\caption{The PixelCNN architecure used for the CIFAR-10 dataset.}
\label{tab:pcnn_arch}
\centering
\begin{tabular}{l}
\hline
3 x 32 x 32 RGB Image \\
\hline
Conv7x7(256) (Mask A) \\
\hline
\textbf{15 Residual Blocks, each using:} \\
\quad ReLU - Conv1x1(256) (Mask B) \\
\quad ReLU - Conv3x3(128) (Mask B) \\
\quad ReLU - Conv1x1(256) (Mask B) \\
\hline
ReLU - Conv1x1(1024) (Mask B) \\
ReLU - Conv1x1(P) (Mask B) \\
\hline
\end{tabular}
\end{table}

For the PixelCNN++ we used, like \citet{salimans2017}, 6 blocks with 5 ResNet layers. Between blocks 1 and 2 and between blocks 2 and 3, strided convolutions are used to downsample the feature maps. Between blocks 4 and 5 and between blocks 5 and 6, transposed strided convolutions are used to upsample the feature maps back to the original size. Shortcut connections are added from block 1 to 6, 2 to 5 and 3 to 4. For more details on the PixelCNN++ architecture see \citet{salimans2017} and their publicly available code\footnote{\url{https://github.com/openai/pixel-cnn}}. 

For models using linear splines we used 256 bins, corresponding to the quantization level. For models using quadratic splines, we used 16 bins as this was found to work well in early experiments. Note that for quadratic splines, the bin locations can also be learned \citep{muller2018}. As a consequence, less bins are typically required than for linear splines.
Finally, for models using DMOL, 10 mixtures were used.

PixelFlow and PixelFlow++ use compositions of a PixelCNN(++) and a Glow-like coupling flow. The coupling flow is a multi-scale architecure \citep{dinh2017} using 2 scales and 8 steps/scale. Each step consists of an affine coupling layer \citep{dinh2017} and an invertible $1 \times 1$ convolution \citep{kingma2018}. The models were trained with variational dequantization \citep{ho2019} and an initial squeezing layer \citep{dinh2017} to increase the number of channels from 3 to 12. The coupling layers are parameterized by DenseNets \citep{huang2017}.

All models were trained for the CIFAR-10 dataset were trained for 500 epochs with a batch size of 16. For all models, except PixelFlow and PixelFlow++, the Adam optimizer \citep{kingma2014} was used with an initial learning rate of $3\cdot 10^{-4}$. For PixelFlow and PixelFlow++, the Adamax optimizer \citep{kingma2014} was used with an initial learning rate of $1\cdot 10^{-3}$. For all models, the learning rate was decayed by $0.5$ at epochs $250, 300, 350, 400, 450$. All models were trained on a single GPU. 
Depending on the type of GPU used, training a stock PixelCNN takes about 30 hours and training a stock PixelCNN++ takes about 10 days.


\section{Interpolation Experiment Details}
\label{app:interp}

In Fig. \ref{fig:interpolations_c10}, latent space interpolations in PixelCNN and PixelCNN++ models are shown. 
In order to obtain these interpolations, we first transform two real images $\vx^{(0)}$ and $\vx^{(1)}$ to the latent regions $f(\mathcal{B}(\vx^{(0)}))$ and $f(\mathcal{B}(\vx^{(1)}))$ and sample according to the uniform base distribution to obtain points $\vz^{(0)}$ and $\vz^{(1)}$ in the latent space. Linearly interpolating in this space does not yield uniform samples. Empirically, we found this to often give blurry, single-color interpolations. To get interpolated points that are valid samples from the uniform distribution, we first further transformed the latent images $\vz^{(0)} \rightarrow \vh^{(0)}$ and $\vz^{(1)} \rightarrow \vh^{(1)}$ using the inverse Gaussian CDF for each dimension. As this is an invertible transformation, we can equivalently consider the base distribution as the isotropic Gaussian. Subsequently, we interpolated according to
\begin{equation*}
    \vh^{(w)} = \frac{w \vh^{(0)} + (1-w) \vh^{(1)}}{\sqrt{w^2 + (1-w)^2}},
\end{equation*}
for $0 \leq w \leq 1$.
This yields a path of equally probable samples under the base distribution, i.e. $\vh^{(w)} \sim \mathrm{N}(0,1)$ for $\vh^{(0)}, \vh^{(1)} \sim \mathrm{N}(0,1)$. Finally, the intermediate latent points $\vh^{(w)}$ are transformed back to samples $\vx^{(w)}$.


\section{Multilayer Subset Flows Experiments}
\label{app:multilayer_subset}

In this section, we investigate multi-layer subset flows. Multi-layer subset flows can tractably transform finite volumes through \emph{multiple} transformations. For \emph{autoregressive subset flows}, this may be achieved by letting all the autoregressive flows share the same autoregressive order.

Denote the intermediate spaces as $\vz_l, l=0,1,...,L$ with $\vy \equiv \vz_0$ and $\vz \equiv \vz_L$. In a layer $l$, we further denote the boundaries of the hyperrectangle for dimension $d$ as $z_{d,l}^{\mathrm{(lower)}}$ and $z_{d,l}^{\mathrm{(upper)}}$. The procedure for computing the likelihood in a multilayer autoregressive subset flow is shown in Algo. \ref{alg:likelihood}, while the procedure for sampling is shown in Algo. \ref{alg:sampling}.

{\centering
\begin{minipage}{.48\linewidth}
\begin{algorithm}[H]
  \caption{Multilayer Subset Flow Likelihood}
  \label{alg:likelihood}
\begin{algorithmic}
  \STATE Observe discrete $\vx$.
  \STATE Define $\vz_0^{\mathrm{(lower)}} = \vx$, $\vz_0^{\mathrm{(upper)}} = \vx+1$.
  \FOR{$l=1$ {\bfseries to} $L$}
        \STATE Compute $\vlambda_l \leftarrow \mathrm{Net}\left(\vz_{l-1}^{\mathrm{(lower)}}\right)$.
        \STATE Transform $\vz_l^{\mathrm{(lower)}} = f_l(\vz_{l-1}^{\mathrm{(lower)}}|\vlambda_l)$.
        \STATE Transform $\vz_l^{\mathrm{(upper)}} = f_l(\vz_{l-1}^{\mathrm{(upper)}}|\vlambda_l)$.
  \ENDFOR
  \STATE $\log P(\vx) = \sum_{d=1}^D \log \left[ z_{L,d}^{\mathrm{(upper)}} - z_{L,d}^{\mathrm{(lower)}} \right]$.
\end{algorithmic}
\end{algorithm}
\end{minipage}
\hspace{0.02\linewidth}
\begin{minipage}{.48\linewidth}
\begin{algorithm}[H]
  \caption{Multilayer Subset Flow Sampling}
  \label{alg:sampling}
\begin{algorithmic}
  \STATE Sample $\vz \sim \prod_{d=1}^D \mathrm{Unif}(z_d|0,1)$.
  \FOR{$d=1$ {\bfseries to} $D$}
      \FOR{$l=1$ {\bfseries to} $L$}
            \STATE Compute $\vlambda_{l,d} \leftarrow \mathrm{Net}\left(\vz_{l-1,1:d-1}^{\mathrm{(lower)}}\right)$.
            \STATE Transform $\vz_{l-1,d} = f_{l,d}^{-1}(z_{l,d}|\vlambda_{l,d})$.
      \ENDFOR
      \STATE Define $y_d = z_{0,d}$.
      \STATE Quantize $y_d$ to obtain $x_d$.
    \ENDFOR
\end{algorithmic}
\end{algorithm}
\end{minipage}
}

\subsection{Multilayer PixelCNN Subset Flows for CIFAR-10}

We will here demonstrate that performance improves with an increasing number of PixelCNNs in a multi-layer subset flow, even when they all share the same autoregressive order.
Due to its relatively lightweight nature compared to more recent autoregressive models, we use the original architecture of \citet{oord2016pixelrnn}. We train 1, 2 and 4 layers of PixelCNN parameterizing either linear or quadratic splines on CIFAR-10. The results can be found in Table \ref{tab:multi_qflow}.  

The single-layer linear spline case corresponds to the original PixelCNN model and the performance reported here matches \citet{oord2016pixelrnn} with 3.14 bits/dim.
We observe that increasing the number of layers improve the fit over the single-layer version. In fact, the 4-layer version is flexible enough to overfit. By countering this using dropout with a rate of 0.2, the test set performance of the 4-layer version greatly improves, yielding the best performing model.
In fact, by simply stacking 4 layers of the original PixelCNN, we outperform the improved model Gated PixelCNN \citep{oord2016gatedpixelcnn}. 
In addition to multiple layers improving performance, we observe that quadratic splines improve performance over linear splines in all cases here. 

\begin{table}[t]
\caption{Multiple layers of subset flows can be used to obtain more expressive models. We here illustrate this by stacking vanilla PixelCNNs parameterizing either linear or quadratic splines in multiple layers for CIFAR-10.
Reported numbers are in bits/dim and training performance in parentheses.}
\label{tab:multi_qflow}
\centering
\begin{tabular}{cccc}
\hline
\textbf{Layers} & \textbf{Dropout} & \textbf{PixelCNN} & \textbf{PixelCNN (Quad.)} \\
\hline
1 & - & 3.14 (3.11) & 3.09 (3.05) \\
\hline
2 & - & 3.07 (2.98) & 3.05 (2.96) \\
\hline
\multirow{2}{*}{4} & - & 3.09 (2.89) & 3.09 (2.88) \\
 & 0.2 & \textbf{3.02} (2.98) & \textbf{3.01} (2.98) \\
\hline
\end{tabular}
\end{table}

\subsection{Multilayer PixelCNN++ Subset Flows for CIFAR-10}
\label{app:pixelcnn_pp}

We here experiment with combining PixelCNN++ \citep{salimans2017} with other models in a multilayer subset flow. We use the original PixelCNN++ architecture and combine this with the original PixelCNN architecture \citep{oord2016pixelrnn} parameterizing quadratic splines with 16 bins. We train this combination in both orders. The PixelCNN architecture here uses a dropout rate of 0.5, matching that of the PixelCNN++ architecture. The results are shown in Table \ref{tab:multi_qflow_pp}. 

We observe some slight improvements from combining PixelCNN++ with other models. Earlier experiments indicated that improving on PixelCNN++ by combining it with more layers of PixelCNN or PixelCNN++ was difficult since this often resulted in severe overfitting. We thus observe only mild improvements over the original model by combining it with other flows. We note that these are the best reported results for CIFAR-10 among models that only use convolutions and do not rely on self-attention. For further improvements, incorporating self-attention in the model would thus probably be of help.

\begin{table}[!ht]
\caption{PixelCNN++ with combined with PixelCNN (Quad.) for the CIFAR-10 dataset.}
\label{tab:multi_qflow_pp}
\centering
\begin{tabular}{cc}
\hline
\textbf{Model} & \textbf{Bits/dim} \\
\hline
PixelCNN++ & 2.924 \\
PixelCNN++ \& PixelCNN (Quad.) & 2.914 \\
PixelCNN (Quad.) \& PixelCNN++ & \textbf{2.906} \\
\hline
\end{tabular}
\end{table}

\subsection{Multilayer PixelCNN Subset Flows for ImageNet 32x32 and 64x64}

In addition to training PixelCNN \citep{oord2016pixelrnn} models in multiple flow layers for CIFAR-10, we here attempt the same for the 32x32 and 64x64 ImageNet datasets. For simplicity, all parameters were kept the same as in the case of CIFAR-10, except that 20 residual blocks were used for ImageNet64 (due to the need of a larger receptive field). The results for PixelCNN, $2\times$PixelCNN and $2\times$PixelCNN (Quad.) are shown in Table \ref{tab:multi_qflow_imagenet}. We observe that performance improves with 2 layers instead of 1 and that swapping linear splines with quadratic splines further improves performance.
Note that the performance of these models are slightly sub-par compared to other results by autoregressive models for these datasets. We attribute this to the fact that most other work use larger models, such as more filters in the convolutional layers, than the simple PixelCNN architectures we used here. Regardless, these results show that stacking more layers also helps for the 32x32 and 64x64 ImageNet datasets.

The ImageNet32 models were trained for 30 epochs with a batch size of 16, while the ImageNet64 models were trained for 20 epochs with a batch size of 8. The Adam optimizer \citep{kingma2014} was used. The learning rate was initially $3\cdot 10^{-4}$ and decayed by $0.5$ at epochs $15, 18, 21, 24, 27$ (for ImageNet32) and epochs $10, 12, 14, 16, 18$ (for ImageNet64).

\begin{table}[!ht]
\caption{Multiple layers of vanilla PixelCNNs for the 32x32 and 64x64 ImageNet datasets.}
\label{tab:multi_qflow_imagenet}
\centering
\begin{tabular}{ccc}
\hline
\textbf{Model} & \textbf{ImageNet32} & \textbf{ImageNet64} \\
\hline
PixelCNN & 3.96 & 3.67 \\
2 $\times$ PixelCNN & 3.91 & 3.63 \\
2 $\times$ PixelCNN (Quad.) & \textbf{3.90} & \textbf{3.61} \\
\hline
\end{tabular}
\end{table}


\section{Samples}
\label{app:samples}

\begin{figure}[!ht]
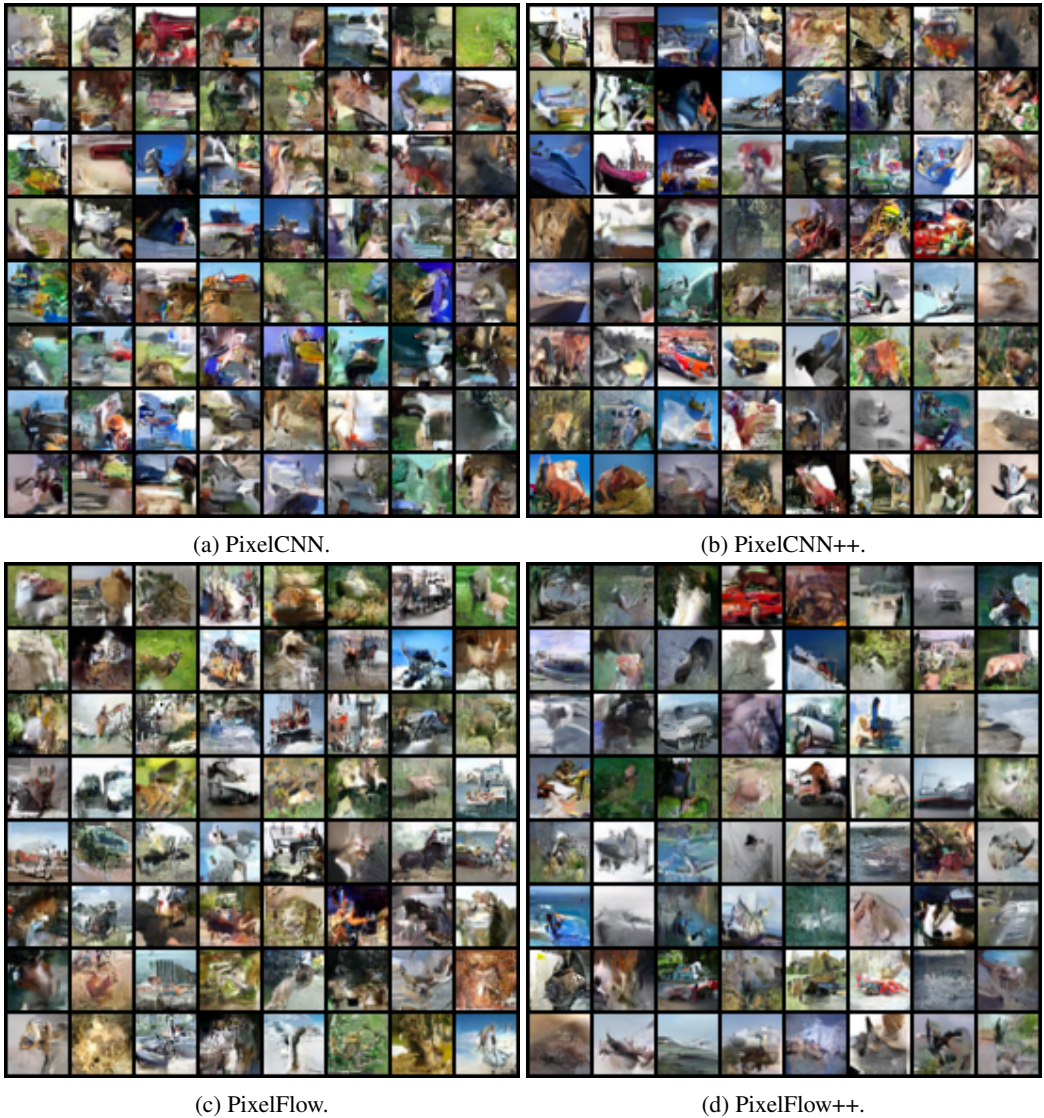

    \centering
    \begin{subfigure}[b]{\sampleswidth}
            \centering
            \includegraphics[width=\textwidth]{figures/samples/pixelcnn.png}
    \subcaption{PixelCNN.}
    \end{subfigure}
    \begin{subfigure}[b]{\sampleswidth}
            \centering
            \includegraphics[width=\textwidth]{figures/samples/pixelcnn_pp.png}
    \subcaption{PixelCNN++.}
    \end{subfigure}
    
    \begin{subfigure}[b]{\sampleswidth}
            \centering
            \includegraphics[width=\textwidth]{figures/samples/pixelflow.png}
    \subcaption{PixelFlow.}
    \end{subfigure}
        \begin{subfigure}[b]{\sampleswidth}
            \centering
            \includegraphics[width=\textwidth]{figures/samples/pixelflow_pp.png}
    \subcaption{PixelFlow++.}
    \end{subfigure}

    \caption{Unconditional samples from PixelCNN-based flow models trained on the CIFAR-10 dataset. 
    The PixelFlow(++) model composes PixelCNN(++) with a Glow-like coupling flow. 
    The perceptual quality appears to improve in the multi-layer flow models compared to the single-layer base models.}
    \label{fig:multilayer_samples}
\end{figure}

See Fig. \ref{fig:multilayer_samples} for samples from PixelCNN, PixelFlow, PixelCNN++ and PixelFlow++.

%% file: main.bbl
\begin{thebibliography}{}

\bibitem[Bengio et~al., 2013]{bengio2013}
Bengio, Y., L{\'{e}}onard, N., and Courville, A.~C. (2013).
\newblock Estimating or propagating gradients through stochastic neurons for
  conditional computation.
\newblock {\em CoRR}, abs/1308.3432.

\bibitem[Burda et~al., 2016]{burda2016}
Burda, Y., Grosse, R.~B., and Salakhutdinov, R. (2016).
\newblock Importance weighted autoencoders.
\newblock In {\em 4th International Conference on Learning Representations}.

\bibitem[Cao et~al., 2019]{cao2019}
Cao, N.~D., Aziz, W., and Titov, I. (2019).
\newblock Block neural autoregressive flow.
\newblock In {\em Proceedings of the Thirty-Fifth Conference on Uncertainty in
  Artificial Intelligence}, page 511.

\bibitem[Chen et~al., 2018]{chen2018}
Chen, X., Mishra, N., Rohaninejad, M., and Abbeel, P. (2018).
\newblock Pixelsnail: An improved autoregressive generative model.
\newblock In {\em Proceedings of the 35th International Conference on Machine
  Learning}, pages 863--871.

\bibitem[Child et~al., 2019]{child2019}
Child, R., Gray, S., Radford, A., and Sutskever, I. (2019).
\newblock Generating long sequences with sparse transformers.
\newblock {\em CoRR}, abs/1904.10509.

\bibitem[Dinh et~al., 2017]{dinh2017}
Dinh, L., Sohl-Dickstein, J., and Bengio, S. (2017).
\newblock {Density estimation using Real NVP}.
\newblock In {\em 5th International Conference on Learning Representations}.

\bibitem[Durkan et~al., 2019]{durkan2019}
Durkan, C., Bekasov, A., Murray, I., and Papamakarios, G. (2019).
\newblock Neural spline flows.
\newblock In {\em Advances in Neural Information Processing Systems}, pages
  7509--7520.

\bibitem[Germain et~al., 2015]{germain2015}
Germain, M., Gregor, K., Murray, I., and Larochelle, H. (2015).
\newblock Made: Masked autoencoder for distribution estimation.
\newblock In {\em Proceedings of the 32nd International Conference on Machine
  Learning}, pages 881--889.

\bibitem[Goodfellow et~al., 2014]{goodfellow2014}
Goodfellow, I., Pouget-Abadie, J., Mirza, M., Xu, B., Warde-Farley, D., Ozair,
  S., Courville, A., and Bengio, Y. (2014).
\newblock Generative adversarial nets.
\newblock In {\em Advances in Neural Information Processing Systems}, pages
  2672--2680.

\bibitem[Grathwohl et~al., 2019]{grathwohl2018}
Grathwohl, W., Chen, R. T.~Q., Bettencourt, J., Sutskever, I., and Duvenaud, D.
  (2019).
\newblock {FFJORD:} free-form continuous dynamics for scalable reversible
  generative models.
\newblock In {\em 7th International Conference on Learning Representations}.

\bibitem[Hinton, 2007]{hinton2007}
Hinton, G.~E. (2007).
\newblock Learning multiple layers of representation.
\newblock {\em Trends in Cognitive Sciences}, 11:428--434.

\bibitem[Hinton et~al., 2006]{hinton2006}
Hinton, G.~E., Osindero, S., and Teh, Y.~W. (2006).
\newblock A fast learning algorithm for deep belief nets.
\newblock {\em Neural Computation}, 18(7):1527--1554.

\bibitem[Ho et~al., 2019]{ho2019}
Ho, J., Chen, X., Srinivas, A., Duan, Y., and Abbeel, P. (2019).
\newblock Flow++: Improving flow-based generative models with variational
  dequantization and architecture design.
\newblock In {\em Proceedings of the 36th International Conference on Machine
  Learning}.

\bibitem[Hoogeboom et~al., 2020]{hoogeboom2020}
Hoogeboom, E., Cohen, T.~S., and Tomczak, J.~M. (2020).
\newblock Learning discrete distributions by dequantization.
\newblock {\em CoRR}, abs/2001.11235.

\bibitem[Hoogeboom et~al., 2019]{hoogeboom2019}
Hoogeboom, E., Peters, J. W.~T., van~den Berg, R., and Welling, M. (2019).
\newblock Integer discrete flows and lossless compression.
\newblock In {\em Advances in Neural Information Processing Systems}, pages
  12134--12144.

\bibitem[Huang et~al., 2018]{huang2018}
Huang, C., Krueger, D., Lacoste, A., and Courville, A.~C. (2018).
\newblock Neural autoregressive flows.
\newblock In {\em Proceedings of the 35th International Conference on Machine
  Learning}, pages 2083--2092.

\bibitem[Huang et~al., 2017]{huang2017}
Huang, G., Liu, Z., van~der Maaten, L., and Weinberger, K.~Q. (2017).
\newblock Densely connected convolutional networks.
\newblock In {\em 2017 {IEEE} Conference on Computer Vision and Pattern
  Recognition}, pages 2261--2269.

\bibitem[Jaini et~al., 2019]{jaini2019}
Jaini, P., Selby, K.~A., and Yu, Y. (2019).
\newblock Sum-of-squares polynomial flow.
\newblock In {\em Proceedings of the 36th International Conference on Machine
  Learning}, pages 3009--3018.

\bibitem[Kalchbrenner et~al., 2017]{kalchbrenner2017}
Kalchbrenner, N., van~den Oord, A., Simonyan, K., Danihelka, I., Vinyals, O.,
  Graves, A., and Kavukcuoglu, K. (2017).
\newblock Video pixel networks.
\newblock In {\em Proceedings of the 34th International Conference on Machine
  Learning}, pages 1771--1779.

\bibitem[Kingma and Ba, 2015]{kingma2014}
Kingma, D.~P. and Ba, J. (2015).
\newblock Adam: {A} method for stochastic optimization.
\newblock In {\em 3rd International Conference on Learning Representations}.

\bibitem[Kingma and Dhariwal, 2018]{kingma2018}
Kingma, D.~P. and Dhariwal, P. (2018).
\newblock Glow: Generative flow with invertible 1x1 convolutions.
\newblock In {\em Advances in Neural Information Processing Systems}, pages
  10236--10245.

\bibitem[Kingma et~al., 2016]{kingma2016}
Kingma, D.~P., Salimans, T., J{\'{o}}zefowicz, R., Chen, X., Sutskever, I., and
  Welling, M. (2016).
\newblock Improving variational autoencoders with inverse autoregressive flow.
\newblock In {\em Advances in Neural Information Processing Systems}, pages
  4736--4744.

\bibitem[Kingma and Welling, 2014]{kingma2013}
Kingma, D.~P. and Welling, M. (2014).
\newblock Auto-encoding variational bayes.
\newblock In {\em 2nd International Conference on Learning Representations}.

\bibitem[Ma et~al., 2019]{ma2019}
Ma, X., Kong, X., Zhang, S., and Hovy, E.~H. (2019).
\newblock Macow: Masked convolutional generative flow.
\newblock In {\em Advances in Neural Information Processing Systems}, pages
  5891--5900.

\bibitem[Menick and Kalchbrenner, 2019]{menick2019}
Menick, J. and Kalchbrenner, N. (2019).
\newblock Generating high fidelity images with subscale pixel networks and
  multidimensional upscaling.
\newblock In {\em 7th International Conference on Learning Representations}.

\bibitem[M{\"{u}}ller et~al., 2019]{muller2018}
M{\"{u}}ller, T., McWilliams, B., Rousselle, F., Gross, M., and Nov{\'{a}}k, J.
  (2019).
\newblock Neural importance sampling.
\newblock {\em {ACM} Trans. Graph.}, 38(5):145:1--145:19.

\bibitem[Oliva et~al., 2018]{oliva2018}
Oliva, J.~B., Dubey, A., Zaheer, M., P{\'{o}}czos, B., Salakhutdinov, R., Xing,
  E.~P., and Schneider, J. (2018).
\newblock Transformation autoregressive networks.
\newblock In {\em Proceedings of the 35th International Conference on Machine
  Learning}, pages 3895--3904.

\bibitem[Papamakarios et~al., 2017]{papamakarios2017}
Papamakarios, G., Murray, I., and Pavlakou, T. (2017).
\newblock Masked autoregressive flow for density estimation.
\newblock In {\em Advances in Neural Information Processing Systems}, pages
  2338--2347.

\bibitem[Parmar et~al., 2018]{parmar2018}
Parmar, N., Vaswani, A., Uszkoreit, J., Kaiser, L., Shazeer, N., Ku, A., and
  Tran, D. (2018).
\newblock Image transformer.
\newblock In {\em Proceedings of the 35th International Conference on Machine
  Learning}, pages 4052--4061.

\bibitem[Rezende and Mohamed, 2015]{rezende2015}
Rezende, D.~J. and Mohamed, S. (2015).
\newblock Variational inference with normalizing flows.
\newblock In {\em Proceedings of the 32nd International Conference on Machine
  Learning}, pages 1530--1538.

\bibitem[Rezende et~al., 2014]{rezende2014}
Rezende, D.~J., Mohamed, S., and Wierstra, D. (2014).
\newblock Stochastic backpropagation and approximate inference in deep
  generative models.
\newblock In {\em Proceedings of the 31th International Conference on Machine
  Learning}, pages 1278--1286.

\bibitem[Salakhutdinov and Hinton, 2009]{salakhutdinov2009}
Salakhutdinov, R. and Hinton, G. (2009).
\newblock Deep boltzmann machines.
\newblock In {\em Proceedings of the International Conference on Artificial
  Intelligence and Statistics}, volume~5, pages 448--455.

\bibitem[Salimans et~al., 2017]{salimans2017}
Salimans, T., Karpathy, A., Chen, X., and Kingma, D.~P. (2017).
\newblock Pixelcnn++: Improving the pixelcnn with discretized logistic mixture
  likelihood and other modifications.
\newblock In {\em 5th International Conference on Learning Representations}.

\bibitem[Song et~al., 2019]{song2019}
Song, Y., Meng, C., and Ermon, S. (2019).
\newblock Mintnet: Building invertible neural networks with masked
  convolutions.
\newblock In {\em Advances in Neural Information Processing Systems}, pages
  11002--11012.

\bibitem[Theis et~al., 2016]{theis2016}
Theis, L., van~den Oord, A., and Bethge, M. (2016).
\newblock A note on the evaluation of generative models.
\newblock In {\em International Conference on Learning Representations}.

\bibitem[Tran et~al., 2019]{tran2019}
Tran, D., Vafa, K., Agrawal, K.~K., Dinh, L., and Poole, B. (2019).
\newblock Discrete flows: Invertible generative models of discrete data.
\newblock In {\em Advances in Neural Information Processing Systems}, pages
  14692--14701.

\bibitem[Uria et~al., 2013]{uria2013}
Uria, B., Murray, I., and Larochelle, H. (2013).
\newblock Rnade: The real-valued neural autoregressive density-estimator.
\newblock In {\em Advances in Neural Information Processing Systems}, pages
  2175--2183.

\bibitem[van~den Oord et~al., 2016a]{oord2016wavenet}
van~den Oord, A., Dieleman, S., Zen, H., Simonyan, K., Vinyals, O., Graves, A.,
  Kalchbrenner, N., Senior, A.~W., and Kavukcuoglu, K. (2016a).
\newblock Wavenet: {A} generative model for raw audio.
\newblock {\em CoRR}, abs/1609.03499.

\bibitem[van~den Oord et~al., 2016b]{oord2016gatedpixelcnn}
van~den Oord, A., Kalchbrenner, N., Espeholt, L., Kavukcuoglu, K., Vinyals, O.,
  and Graves, A. (2016b).
\newblock Conditional image generation with pixelcnn decoders.
\newblock In {\em Advances in Neural Information Processing Systems}, pages
  4790--4798.

\bibitem[van~den Oord et~al., 2016c]{oord2016pixelrnn}
van~den Oord, A., Kalchbrenner, N., and Kavukcuoglu, K. (2016c).
\newblock Pixel recurrent neural networks.
\newblock In {\em Proceedings of the 33nd International Conference on Machine
  Learning}, pages 1747--1756.

\end{thebibliography}
